\documentclass[10pt,twocolumn,letterpaper]{article}

\usepackage[pagenumbers]{wacv} 

\usepackage{booktabs}
\usepackage{colortbl}
\usepackage{subcaption}
\usepackage{graphicx}
\usepackage{textcomp}
\usepackage{multicol}
\usepackage{multirow}
\usepackage{times}
\usepackage{array}
\usepackage{bm}
\usepackage{svg}
\usepackage{dblfloatfix}

\usepackage{pifont}
\newcommand{\cmark}{\ding{51}}

\DeclareMathOperator*{\argmax}{argmax}
\usepackage{amsmath}
\usepackage{amsmath,amssymb,amsfonts}

\definecolor{darkgreen}{RGB}{0,128,0}
\definecolor{darkred}{RGB}{139,0,0}

\definecolor{best}{RGB}{144,238,144}
\definecolor{secondbest}{RGB}{255,255,197}

\definecolor{wacvblue}{rgb}{0.21,0.49,0.74}
\usepackage[pagebackref,breaklinks,colorlinks,allcolors=wacvblue]{hyperref}

\def\wacvPaperID{2266} 
\def\confName{WACV}
\def\confYear{2027}

\title{SUFLECA: Scaling Up Feature Learning for CAD-to-image Alignment}

\author{Saad Ejaz\\
{\small SnT, University of Luxembourg}\\
{\small Luxembourg, Luxembourg}\\
{\tt\small saad.ejaz@uni.lu}
\and
Miguel Fernandez Cortizas\\
{\small SnT, University of Luxembourg}\\
{\small Luxembourg, Luxembourg}\\
{\tt\small miguel.fernandez@uni.lu}
\and
Javier Civera\\
{\small I3A, Universidad de Zaragoza}\\
{\small Zaragoza, Spain}\\
{\tt\small jcivera@unizar.es}
\and
Holger Voos\\
{\small SnT, University of Luxembourg}\\
{\small Luxembourg, Luxembourg}\\
{\tt\small holger.voos@uni.lu}
\and
Jose Luis Sanchez-Lopez\\
{\small SnT, University of Luxembourg}\\
{\small Luxembourg, Luxembourg}\\
{\tt\small joseluis.sanchezlopez@uni.lu}
}

\begin{document}
\maketitle

\begin{abstract}
CAD-to-image alignment aims to estimate an object’s 9D pose (rotation, translation, and anisotropic scale) from a single RGB image, with applications in robotics and augmented reality. Recent zero-shot methods use vision foundation models to match image regions to CAD models; yet their correspondences are typically appearance-driven or unreliable under occlusion or synthetic-to-real domain shift. To address these limitations, we introduce SUFLECA (Scaling Up Feature LEarning for CAD-to-image Alignment), a weakly supervised framework for zero-shot CAD alignment with two key contributions. First, SUFLECA scales up geometry-grounded feature learning from pretrained visual representations through Normalized Object Coordinates (NOCs) supervision on images spanning up to 12 real and synthetic datasets, learning compact geometry-aware features that generalize across domains. Second, we propose a geometrically consistent matching algorithm that establishes reliable CAD-to-image correspondences.
Together, these contributions enable accurate, sub-second alignment per object instance without iterative pose refinement. On ScanNet25k, SUFLECA achieves 32.8\%/42.6\% category/instance accuracy, outperforming the strongest zero-shot baseline by 9.7/12.5 percentage points with a smaller computational footprint, and for the first time on this benchmark, even surpassing existing pose-supervised methods.

Code is available at: \url{https://github.com/snt-arg/SUFLECA}
\end{abstract}
    
\section{Introduction}
\label{sec_intro}

\begin{figure}[t]
    \centering
    \includegraphics[width=0.95\linewidth]{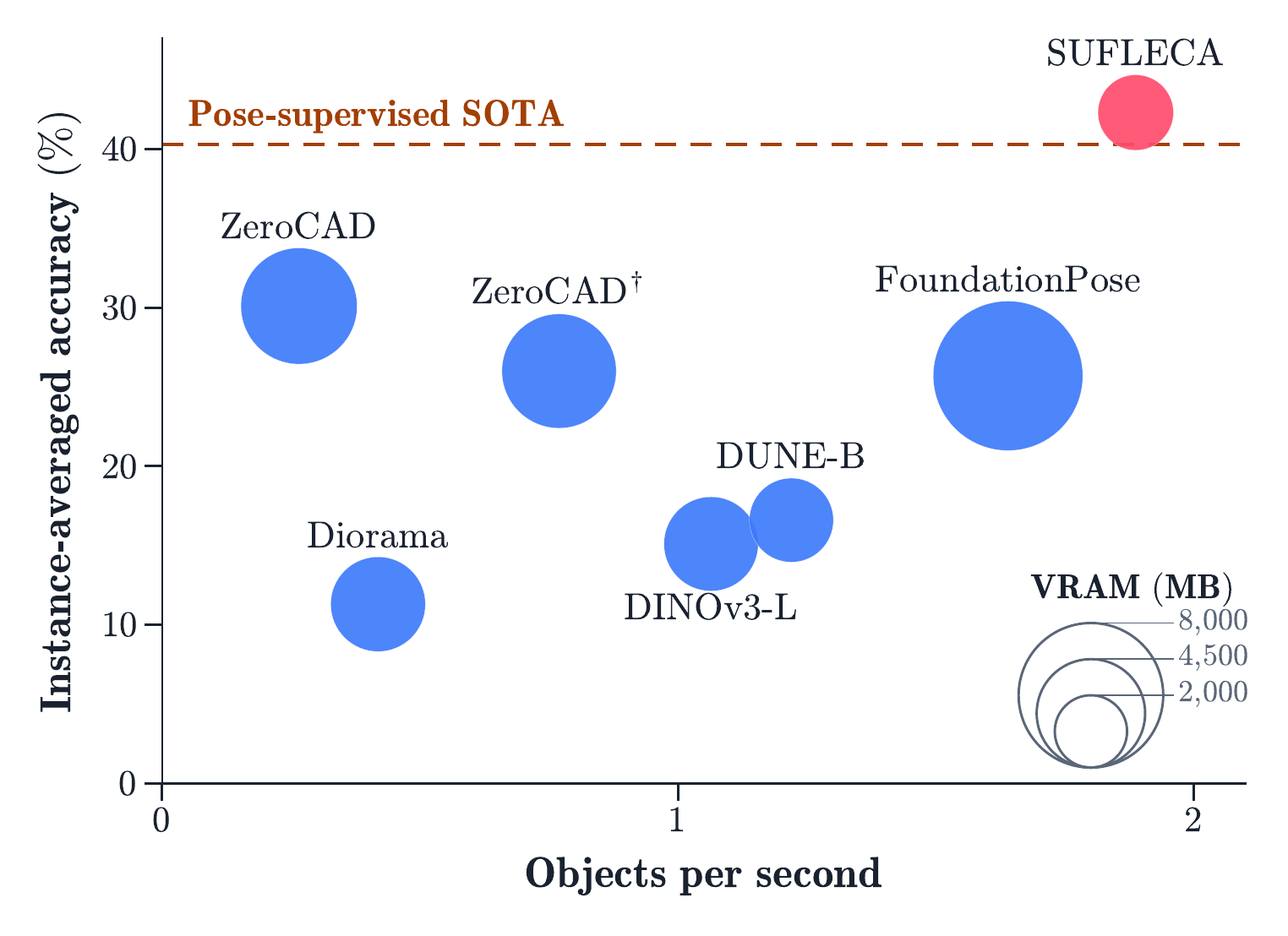}
    \caption{\textbf{Accuracy, runtime, and GPU memory comparison of zero-shot CAD-to-image alignment methods} on the ScanNet25k validation split~\cite{gumeli2022roca}. ZeroCAD$^\dagger$ denotes ZeroCAD~\cite{Arsomngern2025ZeroCAD} without iterative pose refinement. 
    SUFLECA achieves the highest accuracy, surpassing even supervised methods, while attaining the lowest runtime and smallest allocated VRAM among competing zero-shot methods.}
    \label{fig:headline}
\end{figure}

Recovering the 3D pose and shape of objects from images is a fundamental challenge in computer vision, with relevant applications in perception~\cite{shi2023optimal}, manipulation~\cite{collet2009object,tremblay2018corl:dope}, and augmented reality~\cite{hagbi2010shape}.
A practical pipeline typically first retrieves a CAD model that \textit{inexactly} matches the target object and then estimates a 9D alignment (rotation, translation, and anisotropic scale) to register it to the image observation~\cite{gumeli2022roca,Arsomngern2025ZeroCAD}.
The alignment step is typically solved by establishing correspondences between the CAD model and the observed image, followed by robust pose estimation from the resulting matches.

Early works cast this as a supervised learning problem, using end-to-end~\cite{mask2cad,gumeli2022roca} or render-and-compare~\cite{sparc,multisparc} training setups, which lack generalization to out-of-distribution images.
Advances in foundation models for visual perception have enabled zero-shot CAD alignment through pretrained backbones that produce high-dimensional features for correspondence estimation, improving generalization across categories and distributions~\cite{wu2024diorama,antic2025sdfit}.
However, features from visual foundation models are primarily appearance-based and lack grounding in 3D object geometry~\cite{chen2025feat2gs,Arsomngern2025ZeroCAD}. This results in a domain gap, leading to suboptimal alignments.

Weak supervision via synthetic renders of CAD models has been used to align this feature space with the geometry-focused Normalized Object Coordinates (NOCs)~\cite{NOCS,Arsomngern2025ZeroCAD}, yet existing models trained on synthetic and isolated object data struggle to generalize to the clutter, occlusion, and appearance variation present in real scenes.
Beyond training data, nearest-neighbor matching in high-dimensional semantic feature spaces, as used in existing works, 
fails to exploit the rigid geometric structure of the problem.
This produces noisy matches that  require expensive iterative refinement to recover accuracy.

In this paper, we propose \textbf{SUFLECA} (\textbf{S}caling \textbf{U}p \textbf{F}eature \textbf{LE}arning for \textbf{C}AD-to-image \textbf{A}lignment), which addresses these limitations through two key contributions.
First, it scales NOC-aligned feature learning to a large mixture of real and synthetic datasets, substantially improving performance. 
The resulting low-dimensional feature space reduces both matching cost and memory footprint, making it well suited for robotic applications. 
Second, we introduce a correspondence estimation algorithm that replaces conventional semantic-only nearest-neighbor matching with a method that enforces mutuality and geometric consistency.
Together, these contributions enable zero-shot alignment accuracy that surpasses supervised methods while having sub-second runtime, as shown in~\Cref{fig:headline}.

\section{Related Work}
\label{sec_related}

\subsection{Normalized Object Coordinates (NOCs)}
NOCs~\cite{NOCS} represent each object point in a category-level canonical coordinate frame, allowing pose estimation to be formulated as dense coordinate prediction followed by 3D registration.
OmniNOCS~\cite{krishnan2024omninocs} introduced large-scale NOC supervision for pose estimation, although it draws heavily from single-object datasets such as Objectron~\cite{objectron2021}.
Vision backbones implicitly encode 3D awareness~\cite{el2024probing,sariyildiz2025dune}, motivating the use of NOC supervision as an auxiliary signal over frozen features to shape a high-dimensional feature space for correspondence estimation~\cite{Arsomngern2025ZeroCAD}.
However, this remains limited in coverage of occlusions and real-synthetic pairings, which are particularly important in CAD alignment, where a synthetic model is aligned to a real object observation that may be affected by partial occlusion.

\subsection{Supervised CAD-to-image Alignment}
Direct regression methods predict 9D pose in a single feed-forward pass. Total3DUnderstanding~\cite{total_odn} jointly regresses layout, boxes, and meshes, while Mask2CAD~\cite{mask2cad} couples detection with image-shape embeddings.
Render-and-compare methods refine an initial CAD pose by minimizing the difference between the rendered hypothesis and the input image.
3D-RCNN~\cite{kundu20183d} pioneered this idea with differentiable rendering and CAD priors, while SPARC~\cite{sparc} and MultiObj-SPARC~\cite{multisparc} achieve strong results through sparse transformer-based 9D pose updates.
In contrast, indirect methods first estimate correspondences and then recover pose through registration. ROCA~\cite{gumeli2022roca} regresses NOCs from images to establish such correspondences and is trained end-to-end.
More recently, CosCAD~\cite{wen2025coscad} combines cross-modal CAD retrieval with pose alignment by integrating image, CAD, and text features in a shared representation space, improving retrieval and alignment under occlusion.
However, all of these methods rely on direct pose supervision, which limits generalizability and scalability.

\subsection{Zero-shot CAD-to-image Alignment}
A parallel line of work aims to reduce or eliminate direct pose supervision for CAD alignment.
DiffCAD~\cite{diffcad} trains cascaded diffusion models on synthetic data for scale, pose, and shape estimation, while SDFit~\cite{antic2025sdfit} fits a morphable SDF through iterative render-and-compare optimization.
However, both methods require per-category priors and are computationally expensive.
Related category-level approaches~\cite{shi2023optimal,Shaikewitz25-FastShapeAndPose} aim to estimate optimal object pose and shape by progressively pruning outliers to recover the largest set of compatible measurements.
However, these methods still rely on active shape models and keypoint detectors, which are difficult to obtain for most categories.
Among other works, Diorama~\cite{wu2024diorama} uses an additional off-the-shelf model for scale inference~\cite{nguyen2024gigapose} and targets open-world scene-level layout, but its accuracy degrades in real-world scenes with occlusions.
Recently, ZeroCAD~\cite{Arsomngern2025ZeroCAD} employed a NOC-aligned adapter over DINOv2~\cite{oquab2023dinov2} for zero-shot 9D CAD alignment.
However, it is trained exclusively on synthetic single-object scenes, leaving a domain gap, while its nearest-neighbor matching strategy admits many-to-one, geometrically inconsistent correspondences.

\section{SUFLECA}
\label{sec_proposed}

\begin{figure}[t]
    \centering
    \includegraphics[width=\linewidth]{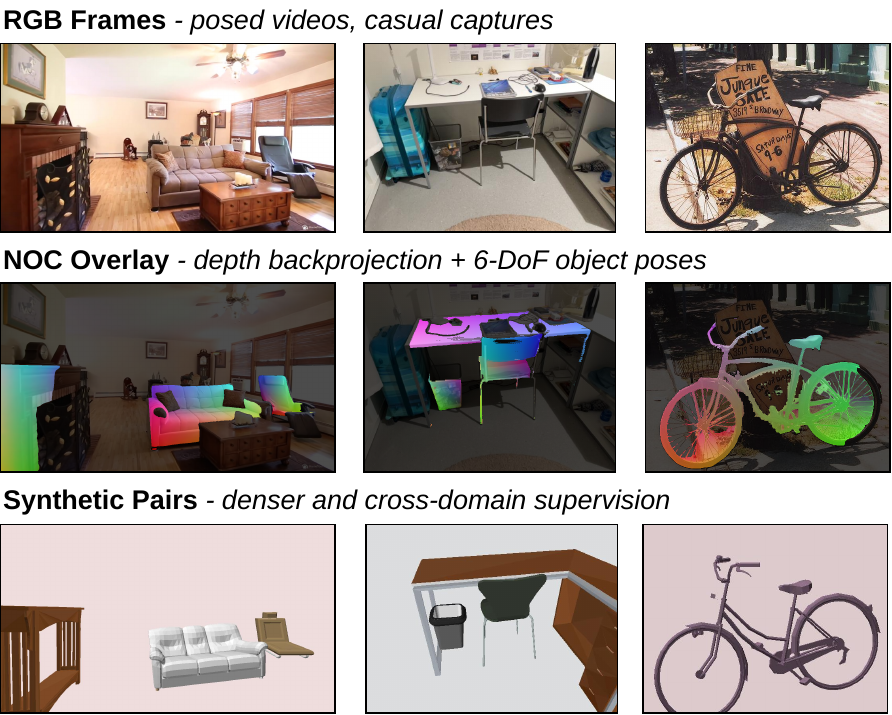}
    \caption{Example frames from the multi-dataset training set. Top row: RGB images. Middle row: ground-truth NOC maps (provided or derived) overlaid on the RGB. Bottom row: synthetic counterparts generated from CAD annotations.}
    \label{fig:data_Examples}
\end{figure}

\subsection{Problem Definition}
Given an RGB image, camera intrinsics, and a CAD model retrieved using an off-the-shelf method, the goal is to estimate a 9D transformation from the CAD coordinate frame $\mathcal{X}_{\mathrm{cad}}$ to the camera coordinate frame $\mathcal{X}_{\mathrm{cam}}$, as follows:
\begin{equation}
    \mathcal{X}_{\mathrm{cam}} = R S \mathcal{X}_{\mathrm{cad}} + t,
\label{eq:problem}
\end{equation}
where $R \in SO(3)$, $S = \mathrm{diag}(s_x, s_y, s_z)$ is an axis-aligned anisotropic scale matrix, and $t \in \mathbb{R}^3$.
We extend this standard formulation to also output a score, $\mathcal{S}_\mathrm{fit} \in \mathbb{R}$, that measures the quality of the estimated alignment, which is useful for ranking alignments or multi-view aggregation.

\subsection{Overall CAD Fitting Pipeline}
SUFLECA follows the standard indirect CAD alignment pipeline used in prior work~\cite{gumeli2022roca,wu2024diorama,Arsomngern2025ZeroCAD,antic2025sdfit}.
\begin{figure*}[t]
    \centering
    \includegraphics[width=0.95\textwidth]{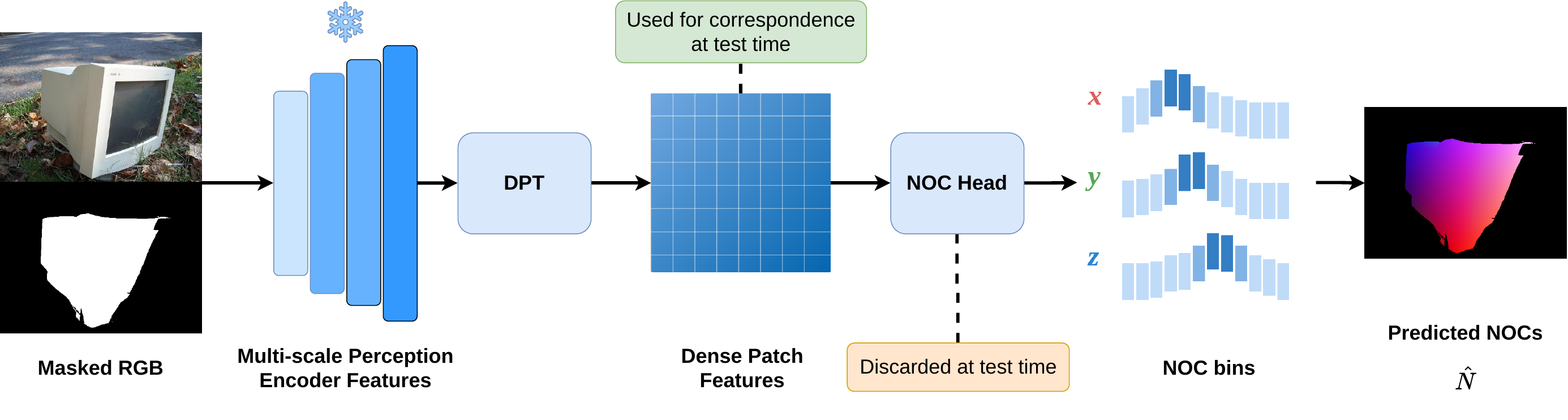}
    \caption{Overview of the feature model architecture. Multi-scale features from a frozen perception encoder are fused and upsampled via a Dense Prediction Transformer (DPT), then decoded by a lightweight binned NOC head. This aligns the high-dimensional DPT features to the NOC space while preserving their discriminative capacity for correspondence estimation.}
    \label{fig:architecture}
\end{figure*}
Given the input image, the pipeline first extracts an object mask and computes per-pixel descriptors.
The same descriptors are also extracted for the candidate CAD model.
Image-to-CAD correspondences are then established in feature space, lifted to 3D using monocular depth estimates, and passed to a RANSAC-based Procrustes solver to estimate the alignment in~\Cref{eq:problem}.
SUFLECA improves two key modules of this pipeline: the feature model, which is learned at scale and unified across real and synthetic data as described in~\Cref{sec:blend}, and the correspondence estimation module, which adds geometric consistency as described in~\Cref{sec:matcher}. 
Throughout this paper, zero-shot refers specifically to the alignment stage, as instance segmentation, depth estimation, and CAD retrieval are treated as fixed upstream components inherited from prior work to ensure comparability.
We additionally evaluate SUFLECA within a fully zero-shot pipeline in the supplementary material.

\begin{table}[t]
    \footnotesize
    \centering
    \resizebox{\columnwidth}{!}{
    \begin{tabular}{
    >{\centering\arraybackslash}p{2.1cm}|
    >{\centering\arraybackslash}p{2.0cm}|
    >{\centering\arraybackslash}p{1.3cm}
    >{\centering\arraybackslash}p{1.4cm}}
        \toprule
        \textbf{RGB frames source} & \textbf{Annotation source} & \textbf{Number of frames} & \textbf{Avg. objects/frame} \\
        \midrule
        Pascal3D+~\cite{everingham2010pascal}        & Included     & 6,\,108   & 1.00  \\
        Objectron~\cite{objectron2021}       & OmniNOCS~\cite{krishnan2024omninocs}     & 7,\,357   & 1.20  \\
        ARKitScenes~\cite{dehghan2021arkitscenes}     & OmniNOCS~\cite{krishnan2024omninocs}     & 52,\,146  & 3.55  \\
        REAL275~\cite{NOCS}         & Included     & 7,\,072   & 5.00  \\
        RealEstate10K~\cite{realestate}   & CAD-Estate~\cite{maninis2023cad}   & 127,\,214  & 2.16  \\
        ScanNet~\cite{dai2017scannet}         & Scannotate~\cite{scannotatepp}   & 117,\,284  & 2.58  \\
        ScanNet++~\cite{yeshwanth2023scannet++}       & Scannotate++~\cite{scannotatepp} & 88,\,198  & 2.96  \\
        Pix3D~\cite{pix3d}           & Included     & 11,\,622   & 1.00  \\
        ObjectNet3D~\cite{xiang2016objectnet3d}     & Included     & 20,\,520  & 1.00  \\
        3D-Front~\cite{3dfront}         & Rendering    & 69,\,578  & 6.43  \\
        Hypersim~\cite{hypersim}        & OmniNOCS~\cite{krishnan2024omninocs}     & 44,\,545  & 13.55 \\
        ShapeNet~\cite{chang2015shapenet}        & Rendering    & 123,207 & 1.00  \\
        \midrule
        \multicolumn{2}{c|}{\textbf{\textit{Overall}}} & \textbf{674,\,851} & \textbf{3.38} \\
        \bottomrule
    \end{tabular}
    }
    \caption{Datasets used for training the feature model. Notably, ScanNet~\cite{dai2017scannet} is omitted from training when evaluating on ScanNet25k~\cite{gumeli2022roca} to maintain a zero-shot setting for this benchmark, leading to a reduced count of 557,\,567 frames.
    }
    \label{tab:datasets}
\end{table}


\subsection{Scaling Unified NOC-aligned Feature Learning}
\label{sec:blend}

\noindent\textbf{Multi-Dataset Setup}\, We adopt a scaling paradigm similar to~\cite{krishnan2024omninocs}, focused primarily on the indoor domain and augmented with synthetic data.
Since correspondences for CAD alignment are established between a real image and a synthetic CAD render, whenever CAD annotations are available, we augment each real image with a synthetic counterpart in which the CAD objects are rendered at approximately the same pose.
The synthetic counterparts are rendered with randomized backgrounds and object textures, with some objects occasionally replaced by visually similar ones.
This yields a larger and more balanced training set with comparable numbers of real and synthetic samples, sharing similar occlusion patterns and scene context, which encourages the model to learn domain-invariant features.
Compared with ~\cite{Arsomngern2025ZeroCAD}, which trains on 300K synthetic images of single objects, our dataset comprises 674K images from 12 datasets (see~\Cref{tab:datasets}) with an average of 3.38 annotated objects per frame across a broader range of categories, providing denser and more diverse supervision.
Furthermore, while most datasets we train on originate from SLAM-based camera trajectories, the inclusion of casual imagery~\cite{pix3d,everingham2010pascal,xiang2016objectnet3d} broadens the data distribution.

\noindent\textbf{NOC Derivation and Verification}\, Most datasets do not provide NOC maps directly, but instead supply object poses in 3D from which NOCs must be derived.
To this end, we render the ground-truth object onto the image to obtain an approximate bounding box, from which a mask is extracted using SAM2~\cite{ravi2024sam2}.
The masked region is then backprojected into 3D using available depth maps or a monocular metric depth estimator~\cite{metricanything2026}, after which NOCs are computed following~\cite{krishnan2024omninocs} (see the middle row in~\Cref{fig:data_Examples}).
To ensure annotation quality, we verify the computed NOCs against the overlaid CAD NOCs and reject object instances whose mean error exceeds a threshold.
Frames for which the total number of valid NOC pixels falls below a threshold are discarded entirely.
For datasets derived from high frame-rate cameras~\cite{realestate,dai2017scannet,yeshwanth2023scannet++}, we subsample frames at regular intervals and additionally enforce a minimum camera pose displacement to ensure sufficient viewpoint diversity.


\noindent\textbf{Model Architecture}\, As shown in~\Cref{fig:architecture}, our model builds upon a pretrained perception encoder, kept frozen during training, that provides rich patch-level features from images.
These patch features are upsampled using a Dense Prediction Transformer (DPT)~\cite{2021dpt}, which fuses multi-scale outputs from different layers of the encoder to produce denser feature maps of dimension $\mathrm{dim}$.
The DPT features are decoded by a lightweight NOC head, i.e., a binned classifier~\cite{NOCS,krishnan2024omninocs} that partitions each coordinate axis ($x$, $y$, $z$) of the NOC space into $m$ uniform bins, estimating each value as the expectation over the bin probabilities.
This supervision encourages the DPT features to be spatially smooth and geometrically meaningful.
The training loss comprises a categorical cross-entropy term between the predicted bin probabilities and the ground-truth bin, and an $\ell_1$ regression term on the expected NOC,
\begin{equation}
    \mathcal{L}_{\mathrm{NOC}} = \mathcal{L}_{\mathrm{CE}} + \lambda \| \hat{\mathcal{N}} - \mathcal{N} \|_1,
\end{equation}
where $\mathcal{L}_{\mathrm{CE}}$ is the cross-entropy loss, $\hat{\mathcal{N}}$ is the predicted NOC value, $\mathcal{N}$ is the ground-truth NOC value, and $\lambda$ is a weighting coefficient.
The NOC head is discarded at test time, and the $\ell_2$-normalized DPT features are used as descriptors for correspondence estimation in SUFLECA.
Optionally, these features are concatenated with the output features of the vision backbone, similar to ~\cite{Arsomngern2025ZeroCAD}, to mitigate forgetting caused by weak supervision over a limited set of categories.
We refer to this variant as SUFLECA-blend.



\subsection{Correspondence Estimation}
\label{sec:matcher}
To align a CAD model $C$ to a masked image $I$,
we start by featurizing the image using the aforementioned feature model $\mathcal{M}$, yielding per-patch features interpolated to the pixel level, $\mathcal{F}_I = \mathcal{M}(I)$.
Unlike prior works that aggregate 2D features across multiple renders of $C$ to obtain 3D features over a mesh~\cite{Arsomngern2025ZeroCAD,antic2025sdfit}, a process that is arbitrary and lossy, we keep the matching task entirely within the 2D domain.
To this end, we maintain a small set of $n$ pre-rendered views of $C$ with precomputed features and point maps.
Perception encoders have been shown to encode global pose information in an image~\cite{wu2024diorama}, which we exploit to select the closest render to the query image prior to estimating correspondences.
To do so, we use a monocular metric depth estimator~\cite{metricanything2026} to lift the masked image pixels into a 3D point cloud $\mathcal{Q}_i$, and apply Farthest Point Sampling (FPS)~\cite{han2023quickfps} to obtain a representative subset $\hat{\mathcal{Q}}_i \subset \mathcal{Q}_i$ of $N$ points for correspondence estimation.
Then, given the set of pre-rendered views $\mathcal{V}_C$ from fixed viewpoints, we select the view $V^*$ that maximizes the mean nearest-neighbor cosine similarity over $\hat{\mathcal{Q}}_i$,
\begin{equation}
    V^* = \argmax_{V \in \mathcal{V}_C} \frac{1}{|\hat{\mathcal{Q}}_i|} \sum_{q \in \hat{\mathcal{Q}}_i} \max_{p \in V} \mathcal{F}_I(q)^\top \mathcal{F}_V(p),
\label{eq:render}
\end{equation}
where $\mathcal{F}_V(p)$ is the precomputed feature at pixel $p$ of view $V$ and $\mathcal{F}_I(q)$ is the feature at pixel $q$ in $I$.

While nearest-neighbor matching suffices for the aforementioned coarse view selection, it is unreliable for correspondence estimation, as it produces geometrically inconsistent and many-to-one matches.
Mutual nearest-neighbor matching reduces these ambiguities but is overly conservative.
We therefore adopt mutual $k$-nearest-neighbors matching, retaining a correspondence only when the render and image descriptors lie within each other's top-$k$ neighbors, with duplicates resolved by greedy assignment to the highest-scoring source match.
However, this only considers the semantic feature space and does not enforce geometric consistency, and since the transformation includes anisotropic scaling, pairwise distance preservation is not directly applicable.
Therefore, similar to \cite{teaser}, we first estimate an approximate scale to normalize the correspondence set before applying pairwise distance preservation.

Concretely, let $\{(\mathbf{p}_i, \mathbf{q}_i)\}_{i=1}^{M}$ be a set of putative correspondences, where $\mathbf{p}_i \in \mathbb{R}^3$ are source points from $V^*$ and $\mathbf{q}_i \in \mathbb{R}^3$ are target points from $I$.
We seek a boolean inlier mask $\mathbf{m} \in \{0,1\}^M$ that retains correspondences geometrically consistent with the unknown 9D transformation.
Denoting by $\mathbf{s}^2 = (s_x^2, s_y^2, s_z^2)^\top$ the squared anisotropic scale vector,
we obtain the following from pairwise differences,
\begin{equation}
    \|\mathbf{q}_i - \mathbf{q}_j\|^2
    = \bm{\Delta}_{ij}^\top \mathrm{diag}(\mathbf{s}^2)\, \bm{\Delta}_{ij}
    = (\mathbf{s}^2)^\top \left(\bm{\Delta}_{ij} \odot \bm{\Delta}_{ij}\right)
\end{equation}
where $\bm{\Delta}_{ij} = \mathbf{p}_i - \mathbf{p}_j$. Normalizing by $\|\bm{\Delta}_{ij}\|^2$ gives the per-pair squared directional scale
\begin{equation}
    \phi_{ij} = \mathbf{f}_{ij}^\top \mathbf{s}^2, \;\;
    \mathbf{f}_{ij} = \frac{\bm{\Delta}_{ij} \odot \bm{\Delta}_{ij}}{\|\bm{\Delta}_{ij}\|^2}, \;\;
    \phi_{ij} = \frac{\|\mathbf{q}_i - \mathbf{q}_j\|^2}{\|\bm{\Delta}_{ij}\|^2}.
\end{equation}
Recovering the full $\mathbf{s}^2$ is ill-conditioned under outliers and incomplete axis coverage, so we estimate only a dominant isotropic scale proxy by taking the histogram mode of the per-pair log-scales $\psi_{ij} = \frac{1}{2}\log \phi_{ij}$ over all $\binom{M}{2}$ pairs, which gives $\hat{\ell} = \mathrm{mode}\{\psi_{ij}\}$ and $\hat{\mathbf{s}}^2 = \exp(2\hat{\ell}) \cdot \mathbf{1}$.
We then build a consensus graph with adjacency $\mathbf{G} \in \{0,1\}^{M\times M}$, comparing each observed squared distance against the value predicted by $\hat{\mathbf{s}}^2$,
\begin{equation}
\begin{split}
    G_{ij} = &\,\mathbf{1}\!\left[\,\left|\frac{\|\mathbf{q}_i - \mathbf{q}_j\|^2 - \hat{d}_{ij}^2}{\hat{d}_{ij}^2}\right|
    < \beta\,\right], \;\; G_{ii} = 0,\\
    \hat{d}_{ij}^2 = &\,(\hat{\mathbf{s}}^2)^\top \left(\bm{\Delta}_{ij} \odot \bm{\Delta}_{ij}\right),
\end{split}
\end{equation}
where $\beta>0$ is the relative inlier threshold.
Under isotropic scaling, correspondences explained by a common transformation form a clique in $\mathbf{G}$. With moderate anisotropy and a sufficiently relaxed $\beta$, a mutually consistent subset of the inliers is expected to form a large clique, which we determine using a computationally bounded version of ~\cite{clique} as:
\begin{equation}
    \mathcal{C}^* = \operatorname*{arg\,max}_{\mathcal{C} \subseteq \{1,\dots,M\}} |\mathcal{C}|
    \quad \text{s.t.} \quad G_{ij} = 1 \;\; \forall\, i \neq j \in \mathcal{C},
\end{equation}
To account for any residual anisotropy along the mis-scaled axes, we use $\mathcal{C}^*$ as a seed and score each correspondence by its agreement with it,
\begin{equation}
    e_i = \frac{1}{|\mathcal{C}^* \setminus \{i\}|}
    \sum_{j \in \mathcal{C}^* \setminus \{i\}} G_{ij}, \qquad
    \mathbf{m} = \mathbf{1}[\mathbf{e} \geq \delta],
\end{equation}
where $e_i = 1$ for $i \in \mathcal{C}^*$ and $\delta \in (0,1]$ gates the minimum agreement with the seed.
The $\mathbf{m}$-masked set with cardinality $\hat{M}$ is passed to a space-partitioning RANSAC~\cite{barath2025superansac} that uses Procrustes registration~\cite{chatrasingh2023generalized} to recover the alignment.

\noindent\textbf{Geometric alignment quality}\, 
Unlike prior works~\cite{gumeli2022roca,Arsomngern2025ZeroCAD}, which use only the semantic detection score as a proxy for alignment quality, SUFLECA estimates $\mathcal{S}_\mathrm{fit}$ from the registration residuals, providing a geometric measure of alignment quality.
The information matrix $\mathbf{\Lambda} \in \mathbb{R}^{9 \times 9}$ is computed from the Jacobians of the registration residuals with respect to the parameter vector $\mathbf{x} = [\delta\boldsymbol{\theta}, \delta\mathbf{t}, \delta\mathbf{s}]$, where $\delta\boldsymbol{\theta}$ is a left rotation perturbation, $\delta\mathbf{t}$ is an additive translation, and $\delta\mathbf{s}$ is an additive perturbation on the diagonal scale.
The residual for each correspondence $i$ is $\mathbf{r}_i = \mathbf{q}_i - RS\mathbf{p}_i - t$, with per-correspondence Jacobian,
\begin{equation}
    \mathbf{J}_i = \begin{bmatrix} [\mathbf{u}_i]_\times & -\mathbf{I}_3 & -R\,\mathrm{diag}(\mathbf{p}_i) \end{bmatrix} \in \mathbb{R}^{3 \times 9},
\end{equation}
where $\mathbf{u}_i = RS\mathbf{p}_i$ is the rotated and scaled source point and $[\cdot]_\times$ denotes the skew-symmetric cross-product matrix.
The information matrix is then
\begin{equation}
    \mathbf{\Lambda} = \frac{1}{\sigma^2}\sum_i \mathbf{J}_i^\top \mathbf{J}_i, \quad \sigma^2 = \frac{\sum_i \|\mathbf{r}_i\|^2}{3\hat{M} - 9},
\label{eq:logdet}
\end{equation}
from which the geometric alignment quality score is computed as $\mathcal{S}_\mathrm{fit}=\log\det(\mathbf{\Lambda})$.
\section{Evaluation}
\label{sec_evaluation}

\begin{table*}[t]
    \footnotesize
    \centering
    \resizebox{\textwidth}{!}{%
    \begin{tabular}{l|
    >{\centering\arraybackslash}p{0.5cm}|
    >{\centering\arraybackslash}p{0.7cm}
    >{\centering\arraybackslash}p{0.7cm}
    >{\centering\arraybackslash}p{0.7cm}
    >{\centering\arraybackslash}p{0.7cm}
    >{\centering\arraybackslash}p{0.7cm}
    >{\centering\arraybackslash}p{0.7cm}
    >{\centering\arraybackslash}p{0.7cm}
    >{\centering\arraybackslash}p{0.7cm}
    >{\centering\arraybackslash}p{0.7cm}|
    >{\centering\arraybackslash}p{0.7cm}
    >{\centering\arraybackslash}p{0.7cm}}
        \toprule
        \multirow{2}{*}{Method} & \multirow{2}{*}{Sup.} & \multicolumn{9}{c|}{Category-specific accuracy (\%)} & \multicolumn{2}{c}{Average (\%)} \\
        & & bathtub & bed & bin & bkshlf & cabinet & chair & display & sofa & table & cat. & inst. \\
        \midrule
        \multicolumn{13}{c}{ScanNet25k~\cite{gumeli2022roca} -- Vid2CAD~\cite{vid2cad} NMS Protocol} \\
        \midrule
        ROCA~\cite{gumeli2022roca}            & \cmark\ & 20.8 & 8.6 & 26.3 & 9.0 & 13.1 & 39.9 & 24.6 & 10.6 & 12.7 & 18.4 & 25.0 \\
        SPARC~\cite{sparc}                    & \cmark\ & \cellcolor{best!60}\textbf{25.8} & 25.7 & 24.6 & 14.2 & 20.8 & 51.5 & 17.8 & 28.3 & 15.4 & 24.9 & 31.8 \\
        MultiObj-SPARC~\cite{multisparc}      & \cmark\ & \cellcolor{best!60}\textbf{25.8} &\cellcolor{best!60}\textbf{34.3} & \cellcolor{best!60}\textbf{44.8} & 17.0 & 19.2 & \cellcolor{secondbest}64.8 & 5.8 & 35.4 & 25.5 & 30.3 & 40.3 \\
        CosCAD~\cite{wen2025coscad}                    & \cmark\ & \cellcolor{secondbest}24.9 & 18.9 & 31.1 & \cellcolor{best!60}\textbf{20.8} & \cellcolor{best!60}\textbf{33.1} & 45.4 & 27.5 & 19.9 & 24.6 & 27.4 & 33.2 \\
        DINOv2-L~\cite{oquab2023dinov2} & -- & 14.2 & 5.7 & 6.0 & 8.5 & 6.9 & 34.3 & 9.9 & 24.8 & 7.2 & 13.1 & 18.8 \\
        DINOv3-L~\cite{simeoni2025dinov3} & -- & 7.5 & 7.1 & 6.5 & 4.7 & 6.5 & 24.7 & 11.0 & 29.2 & 8.7 & 11.8 & 15.1 \\
        DUNE-B~\cite{sariyildiz2025dune} & -- & 6.7 & 2.9 & 8.6 & 2.4 & 6.9 & 28.6 & 12.6 & 23.9 & 10.1 & 11.4 & 16.6 \\
        Diorama~\cite{wu2024diorama} & -- & 6.7 & 1.4 & 19.8 & 3.8 & 4.2 & 16.7 & 11.5 & 9.0 & 6.0 & 8.7 & 11.3 \\
        FoundationPose (9D)~\cite{wen2024foundationpose,Arsomngern2025ZeroCAD} & * & 20.0 & 22.9 & 27.6 & 0.9 & 3.1 & 41.8 & 23.6 & 15.0 & 17.5 & 19.2 & 25.7 \\
        ZeroCAD$^\dagger$ (w/o refine)~\cite{Arsomngern2025ZeroCAD} & * & 16.7 & 10.0 & 10.8 & 15.1 & 7.3 & 46.8 & 16.2 & 31.0 & 10.7 & 18.3 & 26.0 \\
        ZeroCAD~\cite{Arsomngern2025ZeroCAD} & * & 16.7 & 18.6 & 22.8 & 12.7 & 9.2 & 49.3 & 24.1 & 38.1 & 16.5 & 23.1 & 30.1 \\
        SUFLECA (\textit{Ours}) & * & 23.3 & 28.6 & 31.5 & 17.5 & 24.6 & \cellcolor{best!60}\textbf{65.4} & \cellcolor{secondbest}30.9 & 43.4 & \cellcolor{secondbest}29.8 & \cellcolor{secondbest}32.8 & \cellcolor{best!60}\textbf{42.6} \\
        SUFLECA-S (\textit{Ours}) & * & 22.5 & 25.7 & \cellcolor{secondbest}34.9 & \cellcolor{secondbest}17.9 & 22.7 & 60.6 & 24.6 & \cellcolor{secondbest}44.3 & 26.9 & 31.1 & 39.8 \\
        SUFLECA-blend (\textit{Ours}) & * & 23.3 & \cellcolor{secondbest}31.4 & 33.2 & 17.5 & \cellcolor{secondbest}25.8 & 64.3 & \cellcolor{best!60}\textbf{31.4} & \cellcolor{best!60}\textbf{45.1} & \cellcolor{best!60}\textbf{30.2} & \cellcolor{best!60}\textbf{33.6} & \cellcolor{best!60}\textbf{42.6} \\
        \midrule
        \multicolumn{13}{c}{DiffCAD~\cite{diffcad} Split} \\
        \midrule
        \textcolor{gray}{DiffCAD (GT)~\cite{diffcad}} & \textcolor{gray}{*} & \textcolor{gray}{--} & \textcolor{gray}{27.1} & \textcolor{gray}{--} & \textcolor{gray}{24.4} & \textcolor{gray}{33.0} & \textcolor{gray}{65.9} & \textcolor{gray}{--} & \textcolor{gray}{46.3} & \textcolor{gray}{18.3} & \textcolor{gray}{35.8} & \textcolor{gray}{41.9} \\
        DINOv3-L~\cite{simeoni2025dinov3} & -- & -- & 5.2 & -- & 9.2 & 12.5 & 25.1 & -- & 23.8 & 5.9 & 13.6 & 16.0 \\
        DUNE-B~\cite{sariyildiz2025dune} & -- & -- & 3.2 & -- & 9.2 & 13.1 & 26.0 & -- & 23.4 & 8.7 & 14.0 & 17.0 \\
        Diorama~\cite{wu2024diorama} & -- & -- & 0.7 & -- & 5.4 & 13.1 & 33.5 & -- & 6.5 & 12.2 & 11.9 & 18.4 \\
        DiffCAD~\cite{diffcad} & * & -- & 7.7 & -- & 7.6 & 10.4 & 31.1 & -- & 15.3 & 4.3 & 12.7 & 16.7 \\
        ZeroCAD~\cite{Arsomngern2025ZeroCAD} & * & -- & 2.8 & -- & 8.2 & 12.0 & 41.3 & -- & 21.4 & 9.8 & 15.9 & 20.9 \\
        SUFLECA (\textit{Ours}) & * & -- & \cellcolor{secondbest}21.3 & -- & \cellcolor{secondbest}23.9 & \cellcolor{secondbest}34.7 & \cellcolor{secondbest}70.5 & -- & \cellcolor{secondbest}42.9 & \cellcolor{secondbest}25.9 & \cellcolor{secondbest}36.5 & \cellcolor{secondbest}45.0 \\
        SUFLECA-S (\textit{Ours}) & * & -- & \cellcolor{secondbest}21.3 & -- & \cellcolor{best!60}\textbf{25.0} & 31.7 & 68.3 & -- & 41.8 & 22.1 & 35.0 & 42.8 \\
        SUFLECA-blend (\textit{Ours}) & * & -- & \cellcolor{best!60}\textbf{21.9} & -- & 21.7 & \cellcolor{best!60}\textbf{32.0} & \cellcolor{best!60}\textbf{72.0} & -- & \cellcolor{best!60}\textbf{44.4} & \cellcolor{best!60}\textbf{27.6} & \cellcolor{best!60}\textbf{36.6} & \cellcolor{best!60}\textbf{45.7} \\
        \bottomrule
    \end{tabular}}
    \caption{Evaluation on ScanNet25k NMS and DiffCAD split. We report per-category and per-instance average accuracies for $20\,\mathrm{cm}$/$20^\circ$/$20\%$ thresholds. \cmark\ denotes full 9D pose supervision; * denotes weak supervision; and -- stands for unsupervised methods. Both * and -- operate in the zero-shot setting. \colorbox{best!60}{\textbf{Bold}} and \colorbox{secondbest}{shaded} indicate best and second-best results respectively.}
    \label{tab:main}
\end{table*}

\subsection{Implementation Details}
\label{sec:impl}
We use DUNE-B~\cite{sariyildiz2025dune} as the frozen perception encoder and produce DPT patch features with $\mathrm{dim}=384$.
These are fed to the NOC head, which consists of two shallow convolutional layers and classifies coordinates into $m=64$ bins.
We also train a smaller variant, SUFLECA-S, which uses DUNE-S~\cite{sariyildiz2025dune} as the encoder with $\mathrm{dim}=256$ and $m=50$.


\subsection{Methodology}
We evaluate SUFLECA across complementary settings.
ScanNet25k~\cite{gumeli2022roca} serves as the main benchmark for alignment accuracy, while CO3D~\cite{reizenstein21co3d} is adapted to assess generalization to categories not observed during weak supervision (\Cref{sec:results}).
We additionally use ReplicaCAD~\cite{szot2021habitat} as a controlled synthetic setting to analyze the contribution of real and synthetic training data.
We further include ablation studies to isolate the contributions of our design choices (\Cref{sec:ablation}), and perform an efficiency analysis (\Cref{sec:timing}).
We also report results with the smaller SUFLECA-S variant (from \Cref{sec:impl}) and the SUFLECA-blend variant (from \Cref{sec:blend}) of our method.

\begin{table*}[t]
    \footnotesize
    \centering
    \resizebox{\linewidth}{!}{%
    \begin{tabular}{
    >{\arraybackslash}p{3.0cm}|
    >{\centering\arraybackslash}p{1.25cm}
    >{\centering\arraybackslash}p{1.25cm}
    >{\centering\arraybackslash}p{1.25cm}
    >{\centering\arraybackslash}p{1.65cm}|
    >{\centering\arraybackslash}p{1.25cm}
    >{\centering\arraybackslash}p{1.25cm}
    >{\centering\arraybackslash}p{1.25cm}
    >{\centering\arraybackslash}p{1.65cm}}
        \toprule
        \multirow{2}{*}{Method} & \multicolumn{4}{c|}{Seen categories (CAD retrieval accuracy: 77.5\%)} & \multicolumn{4}{c}{Unseen categories (CAD retrieval accuracy: 52.5\%)} \\
        & 3D-IoU ↑ & ICP-Rot ↓ & ADD-S ↓ & ADD-S@0.1↑ & 3D-IoU ↑ & ICP-Rot ↓ & ADD-S ↓ & ADD-S@0.1↑ \\
        \midrule
        DINOv3-L~\cite{simeoni2025dinov3} & 36.21 & 27.32 & 6.54 & 71.50 & 8.77 & 40.44 & 10.31 & 46.50 \\
        DUNE-B~\cite{sariyildiz2025dune} & 23.40 & 35.81 & 8.48 & 63.50 & 4.46 & 44.44 & 11.84 & 34.50 \\
        Diorama~\cite{wu2024diorama} & 39.41 & 21.06 & 5.79 & 82.00 & 39.12 & 14.54 & 6.33 & \cellcolor{best!60}\textbf{81.50} \\
        SUFLECA (\textit{Ours}) & \cellcolor{secondbest}62.30 & \cellcolor{secondbest}9.52 & 3.77 & \cellcolor{best!60}\textbf{86.00} & \cellcolor{secondbest}48.77 & \cellcolor{secondbest}14.53 & \cellcolor{secondbest}5.73 & \cellcolor{secondbest}79.25 \\
        SUFLECA-S (\textit{Ours}) & 62.23 & 10.51 & \cellcolor{secondbest}3.75 & \cellcolor{best!60}\textbf{86.00} & 48.14 & 17.00 & 6.31 & 74.75 \\
        SUFLECA-blend (\textit{Ours}) & \cellcolor{best!60}\textbf{65.10} & \cellcolor{best!60}\textbf{9.06} & \cellcolor{best!60}\textbf{3.74} & \cellcolor{secondbest}85.50 & \cellcolor{best!60}\textbf{49.18} & \cellcolor{best!60}\textbf{13.20} & \cellcolor{best!60}\textbf{5.45} & \cellcolor{best!60}\textbf{81.50} \\
        \bottomrule
    \end{tabular}}
    \caption{Evaluation on 600 images from selected CO3D~\cite{reizenstein21co3d} sequences.
    We report the mean of the per-category median metrics.
    Seen categories are \textit{chair} and \textit{couch}; unseen categories are \textit{toaster}, \textit{hairdryer}, \textit{microwave}, and \textit{suitcase}.
    3D-IoU and ADD-S@0.1 are reported in \%, ICP-Rot in degrees, and ADD-S is scaled by 100.}
    \label{tab:co3d}
\end{table*}

\subsection{Datasets}
\noindent\textbf{ScanNet25k}~\cite{gumeli2022roca} is derived from the ScanNet~\cite{dai2017scannet} dataset with Scan2CAD annotations~\cite{avetisyan2019scan2cad}. An alignment is considered correct when translation, rotation, and scale errors are below $20\,\mathrm{cm}$, $20^\circ$, and $20\%$, simultaneously.
For the NMS-based evaluation protocol~\cite{vid2cad}, we rank detections by a \textit{mixed-rank} score that multiplies the per-scene percentile ranks of the ROCA detector confidence and $\mathcal{S}_{\mathrm{fit}}$.
To maintain a zero-shot setting on this dataset, we exclude ScanNet from the training data used for SUFLECA's feature models.
Following~\cite{Arsomngern2025ZeroCAD}, all methods use ROCA bounding boxes and CAD retrievals~\cite{gumeli2022roca}, SAM2 masks~\cite{ravi2024sam2}, and depth from a monocular metric depth model~\cite{metricanything2026} fine-tuned on the ScanNet25k training split.

\noindent The \textbf{DiffCAD split}~\cite{diffcad} covers six of the ScanNet25k object categories, evaluated without NMS and using non-fine-tuned depth.
This split was introduced by DiffCAD~\cite{diffcad}, whose priors are available only for six categories.

\noindent\textbf{CO3D}~\cite{reizenstein21co3d} is an object-centric dataset that we adapt for the inexact CAD fitting setting.
We convert its COLMAP~\cite{schoenberger2016sfm} point clouds to metric scale by selecting the depth scale with the lowest error against monocular depth estimates.
We also simulate occlusions by placing a rectangular occluder within the GroundingDINO~\cite{liu2024grounding} bounding box.
For CAD retrieval, we use SAM3D~\cite{sam3dteam2025sam3d3dfyimages} to generate category-level CAD pools and retrieve candidates using a zero-shot strategy adapted from OSCAR~\cite{pulli2026oscar}, described in the supplementary material.
We divide the categories into two groups: seen categories (\textit{chair}, \textit{couch}), which appear in the weakly supervised training data, and unseen categories (\textit{toaster}, \textit{hairdryer}, \textit{microwave}, \textit{suitcase}), which, along with synonymous category labels, are excluded from all constituent training datasets and are used to evaluate generalization performance.
We sample 100 frames per category across 208 unique objects.
Since CO3D lacks ground-truth poses, we use the geometry-based 3D-IoU, ICP-Rot, ADD-S, and ADD-S@0.1 metrics from~\cite{sam3dteam2025sam3d3dfyimages}.

\noindent\textbf{ReplicaCAD}~\cite{szot2021habitat} is a synthetic indoor dataset in which each object corresponds to an exact CAD asset with a known 9D pose.
We render 3,457 frames across 90 scene configurations, covering 1,087 object instances from 12 categories, and perform retrieval over all 92 assets.
For category- and instance-averaged accuracy, we use the same 20\,cm/$20^\circ$/20\% alignment thresholds as ScanNet25k, and additionally report the geometry-based 3D-IoU, ICP-Rot, and ADD-S metrics.
This dataset is used only to analyze the composition of the training data used by SUFLECA.

\subsection{Results}
\label{sec:results}
\Cref{tab:main} reports alignment accuracy on ScanNet25k and the respective DiffCAD split. 
SUFLECA achieves $32.8\%$ category-averaged and $42.6\%$ instance-averaged accuracy using the NMS protocol, surpassing the best zero-shot method, ZeroCAD~\cite{Arsomngern2025ZeroCAD} by $9.7$ and $12.5$ percentage points respectively, and outperforming FoundationPose~\cite{wen2024foundationpose} by a larger margin.
Notably, SUFLECA also outperforms the state-of-the-art MultiObj-SPARC~\cite{multisparc} despite the latter being trained with 9D pose supervision on ScanNet.
On the DiffCAD split, competing zero-shot methods perform poorly because alignment errors are not suppressed through NMS.
In contrast, SUFLECA more than doubles the performance of ZeroCAD~\cite{Arsomngern2025ZeroCAD}, achieving gains of $20.6$ and $24.1$ percentage points in category- and instance-averaged accuracy, respectively.
Moreover, SUFLECA even outperforms the oracle variant of DiffCAD~\cite{diffcad}, which uses ground-truth pose to select the best result from eight hypotheses.
Notably, the other SUFLECA variants also achieve competitive performance, outperforming competing zero-shot methods by a large margin and remaining comparable to supervised methods.
SUFLECA-S attains lower overall accuracy than the base model, but still substantially outperforms other zero-shot methods, highlighting the effectiveness of its compact feature space.
SUFLECA-blend performs on par with SUFLECA since the added DUNE-B features primarily improve generalization, which is less critical for the common object categories in ScanNet25k.

\begin{figure}[t]
    \centering
    \includegraphics[width=\linewidth]{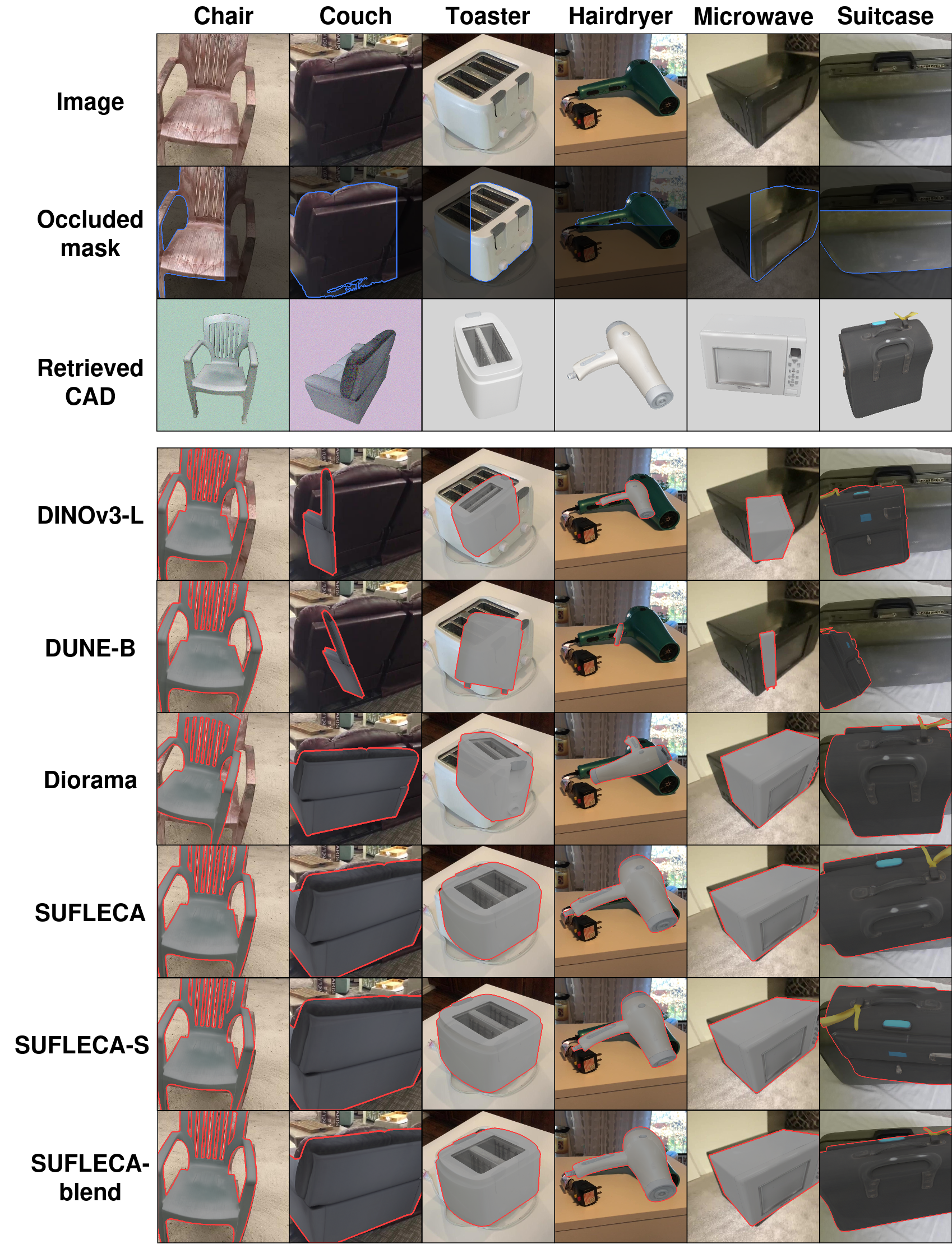}
    \caption{Qualitative comparison of zero-shot methods in aligning an inexact retrieved CAD on masked images from CO3D. The top grid is the input to all methods.}
    \label{fig:qual}
\end{figure}

On CO3D, results are obtained under occluded masks and inexact CADs,
posing significant front--back and left--right geometric ambiguities, as shown in~\Cref{fig:qual}. 
In seen categories, all SUFLECA variants perform comparably and outperform competing methods, highlighting the effectiveness of our weak supervision. For unseen categories, SUFLECA-blend is generally the strongest, as it compensates for the limited supervision coverage for uncommon objects at the cost of a larger feature representation.

\subsection{Training-data composition}

\begin{table*}[t]
\footnotesize
\centering
\resizebox{\linewidth}{!}{
\begin{tabular}{
>{\arraybackslash}p{3.5cm}|
>{\centering\arraybackslash}p{0.9cm}|
>{\centering\arraybackslash}p{1.25cm}
>{\centering\arraybackslash}p{1.25cm}
>{\centering\arraybackslash}p{1.25cm}
>{\centering\arraybackslash}p{1.25cm}
>{\centering\arraybackslash}p{1.25cm}|
>{\centering\arraybackslash}p{1.25cm}
>{\centering\arraybackslash}p{1.25cm}}
\toprule
\multirow{3}{*}{Data composition} & \multirow{3}{*}{\# frames} & \multicolumn{5}{c|}{ReplicaCAD~\cite{szot2021habitat}} & \multicolumn{2}{c}{ScanNet25k~\cite{gumeli2022roca}} \\
&& 3D-IoU & ICP-Rot& ADD-S & cat. & inst. & cat. & inst. \\
&& (\%) ↑ & ($^\circ$) ↓ & ($\times100$) ↓ & (\%) ↑ & (\%) ↑ & (\%) ↑ & (\%) ↑ \\
\midrule
Real + synthetic (\textit{Ours}) & 557k & \cellcolor{best!60}{\textbf{82.26}} & \cellcolor{secondbest}3.98 & \cellcolor{secondbest}1.59 & \cellcolor{secondbest}44.2 & \cellcolor{secondbest}48.7 & \cellcolor{best!60}{\textbf{32.8}} & \cellcolor{best!60}{\textbf{42.6}} \\
Without synthetic counterparts & 430k & 80.50 & 4.37 & 1.62 & 43.2 & 48.0 & \cellcolor{best!60}{\textbf{32.8}} & \cellcolor{secondbest}42.3 \\
Synthetic data only & 422k & \cellcolor{secondbest}82.14 & \cellcolor{best!60}{\textbf{3.88}} & \cellcolor{best!60}{\textbf{1.52}} & \cellcolor{best!60}{\textbf{45.2}} & \cellcolor{best!60}{\textbf{50.9}} & \cellcolor{secondbest}30.8 & 40.6 \\
\bottomrule
\end{tabular}}
\caption{
Effect of training-data composition on alignment performance on synthetic
ReplicaCAD~\cite{szot2021habitat} and real-world
ScanNet25k~\cite{gumeli2022roca}, using the base SUFLECA model.
ScanNet is excluded from all configurations to preserve the zero-shot
ScanNet25k evaluation setting.
}
\label{tab:counterpart}
\end{table*}

To analyze how real and synthetic supervision contribute to the learned representation, we evaluate SUFLECA under different training-data compositions on ReplicaCAD~\cite{szot2021habitat} and ScanNet25k. As shown in Table~\ref{tab:counterpart}, synthetic-only training performs best on most metrics on the synthetic ReplicaCAD but reduces ScanNet25k accuracy to $30.8$/$40.6$. Adding real training imagery improves real-world performance while retaining strong synthetic-domain accuracy. Augmenting the real imagery with synthetic counterparts provides further modest improvements, particularly on ReplicaCAD, while yielding similar performance on ScanNet25k. Overall, the combined real and synthetic training mixture provides the most balanced performance across both domains, supporting their complementary roles in SUFLECA's feature learning. Notably, 
some of the observed improvements may also be attributable to the larger training set rather than data composition alone.

\subsection{Ablation Study}
\label{sec:ablation}
We ablate the main design choices by reporting alignment accuracy on the ScanNet25k validation set in~\Cref{tab:ablation}.

\noindent\textbf{Feature backbone}\,
Replacing our features with those from DINOv3-L~\cite{simeoni2025dinov3} or DUNE-B~\cite{sariyildiz2025dune} substantially reduces alignment accuracy, demonstrating the effectiveness of SUFLECA features.
However, both backbones perform better with our correspondence estimation than in the corresponding results in~\Cref{tab:main}, validating the contributions of both the learned features and the matching pipeline.

\noindent\textbf{Training data}\,
Including ScanNet-derived images during training improves performance from $32.8/42.6$ to $34.4/43.3$.
The modest $1.6/0.7$ percentage-point gain indicates that additional in-distribution data is beneficial, while SUFLECA already generalizes well without it.

\noindent\textbf{Correspondence estimation}\,
Nearest-neighbor matching reduces accuracy to $26.3/34.2$, while enforcing mutual consistency improves it to $30.7/39.4$.
Mutual $k$-nearest-neighbor matching further reaches $32.0/41.0$, with the full geometric-consistency filtering achieving the best performance.
This progression shows that both mutual matching and geometric filtering contribute to accurate alignment.

\noindent\textbf{NMS scoring}\,
Our mixed-rank score, combining the per-scene percentile ranks of the ROCA detector confidence and $\mathcal{S}_{\mathrm{fit}}$, achieves the best accuracy of $32.8/42.6$.
Using $\mathcal{S}_{\mathrm{fit}}$ or the detector score alone reduces performance to $31.4/40.1$ and $30.2/39.7$, respectively.
This shows that semantic detector confidence and geometric alignment quality provide complementary ranking signals.

\begin{table}[t]
    \footnotesize
    \centering
    \resizebox{\linewidth}{!}{%
    \begin{tabular}{
    >{\raggedright\arraybackslash}p{5.9cm}
    >{\centering\arraybackslash}p{1.0cm}
    >{\centering\arraybackslash}p{1.0cm}}
        \toprule
        Variant & cat. & inst.\\
        \midrule
        SUFLECA \textit{(Ours)} --- 384~$\mathrm{dim}$, 102M params. & \textbf{32.8} & \textbf{42.6} \\
        \midrule
        \multicolumn{3}{l}{\textit{Feature backbone (with our correspondence estimation)}} \\
        \quad DINOv3-L (1024~$\mathrm{dim}$, 300M)~\cite{simeoni2025dinov3} &  20.9\textcolor{darkred}{↓} & 25.7\textcolor{darkred}{↓} \\
        \quad DUNE-B (768~$\mathrm{dim}$, 86M)~\cite{sariyildiz2025dune} &  20.4\textcolor{darkred}{↓} & 29.6\textcolor{darkred}{↓} \\
        \multicolumn{3}{l}{\textit{Training data (baseline is 557k images)}} \\
        \quad ScanNet included (674k images) & 34.4\textcolor{darkgreen}{↑} & 43.3\textcolor{darkgreen}{↑} \\
        \multicolumn{3}{l}{\textit{Correspondence estimation (with our features)}} \\
        \quad Nearest neighbor~\cite{gumeli2022roca,Arsomngern2025ZeroCAD} & 26.3\textcolor{darkred}{↓} & 34.2\textcolor{darkred}{↓} \\
        \quad Mutual nearest-neighbor & 30.7\textcolor{darkred}{↓} & 39.4\textcolor{darkred}{↓} \\
        \quad Mutual $k$-nearest-neighbors w/o our filtering & 32.0\textcolor{darkred}{↓} & 41.0\textcolor{darkred}{↓} \\
        \multicolumn{3}{l}{\textit{NMS scoring}} \\
        \quad $\log\det(\mathbf{\Lambda})$ --~\Cref{eq:logdet} & 31.4\textcolor{darkred}{↓} & 40.1\textcolor{darkred}{↓}\\
        \quad ROCA retrieval score~\cite{gumeli2022roca} & 30.2\textcolor{darkred}{↓} & 39.7\textcolor{darkred}{↓} \\

        \bottomrule
    \end{tabular}}
    \caption{Ablation study on the ScanNet25k validation set, isolating one component per group while keeping the rest fixed.}
    \label{tab:ablation}
\end{table}


\subsection{Efficiency Analysis}
\label{sec:timing}
\noindent\textbf{Setup}\, We compare the CAD alignment runtime and peak VRAM usage of different zero-shot methods, with the results reported in~\Cref{tab:timing}.
We exclude detection, segmentation, depth estimation, and retrieval, as these stages are consistent across all methods.
Moreover, since ZeroCAD~\cite{Arsomngern2025ZeroCAD} is not open-source, we implement an approximate runtime-only proxy referred to as ZeroCAD*.


\begin{table}[t]
    \centering
    \footnotesize
    \resizebox{0.80\linewidth}{!}{%
    \begin{tabular}{
    >{\raggedright\arraybackslash}p{3.3cm}|
    >{\centering\arraybackslash}p{0.6cm}|
    >{\centering\arraybackslash}p{1.0cm}
    >{\centering\arraybackslash}p{1.0cm}}
        \toprule
        \multirow{2}{*}{Method} & \multirow{2}{*}{$\mathrm{dim}$} & VRAM & Per-inst.\\
        &&(MB) ↓ &(s) ↓ \\
        \midrule
        DINOv3-L~\cite{simeoni2025dinov3} & 1024 & 3394 & 0.94 \\
        DUNE-B~\cite{sariyildiz2025dune} & 768 & 2696 & 0.82 \\
        Diorama~\cite{wu2024diorama} & 1024 & 3426 & 2.39 \\
        DiffCAD~\cite{diffcad} & -- & 3146 & 13.10 \\
        FoundationPose~\cite{wen2024foundationpose} & -- & 8544 & 0.61  \\
        ZeroCAD$^{*\dagger}$ (w/o refine)~\cite{Arsomngern2025ZeroCAD} & 2048 & 4988 & 1.30 \\
        ZeroCAD$^*$~\cite{Arsomngern2025ZeroCAD} & 2048 & 5158 & 3.77\\
        SUFLECA (\textit{Ours}) & \cellcolor{secondbest}384 & \cellcolor{secondbest}2178 & \cellcolor{secondbest}0.53\\
        SUFLECA-S (\textit{Ours}) & \cellcolor{best!60}\textbf{256} & \cellcolor{best!60}\textbf{1556} & \cellcolor{best!60}\textbf{0.49}\\
        SUFLECA-blend (\textit{Ours}) & 1152 & 3972 & 0.76\\
        \midrule
        \multicolumn{4}{c}{SUFLECA -- CAD alignment timing by component (ms)} \\
        \midrule
        \multicolumn{3}{l}{Pre-processing, featurization, and post-processing} & 86.6 \\
        \multicolumn{3}{l}{Target render selection} & 43.1 \\
        \multicolumn{3}{l}{Correspondence estimation and filtering} & 68.7 \\
        \multicolumn{3}{l}{3D registration w/ RANSAC (300k iterations)} & 332.0 \\
        \bottomrule
    \end{tabular}
    }
    \caption{Peak GPU VRAM and runtime for CAD-to-image alignment across zero-shot methods. 
    ZeroCAD$^*$ denotes a runtime-only proxy of~\cite{Arsomngern2025ZeroCAD}.
    }
    \label{tab:timing}
\end{table}

\noindent\textbf{Discussion}\, SUFLECA achieves lower runtime and VRAM usage than competing methods owing to its compact feature representation (uses between 2.0× and 5.3× fewer dimensions than those of competing methods), and the absence of iterative pose refinement stages as used in~\cite{Arsomngern2025ZeroCAD,wen2024foundationpose}.
The smaller SUFLECA-S further improves runtime and VRAM usage without a significant sacrifice in accuracy (\Cref{tab:main}), while SUFLECA-blend improves generalization (\Cref{tab:co3d}) at the cost of increased runtime and VRAM usage.
Notably, most of the runtime in SUFLECA is attributed to RANSAC, enabling an accuracy-speed trade-off by adjusting the maximum iteration count.

\section{Limitations}
SUFLECA is trained primarily on indoor scenes with mostly furniture-related objects, limiting generalization to uncommon categories.
It also assumes rigid objects without considering articulation or deformation during correspondence estimation.
For example, a mismatch between a CAD template and an articulated object's observed configuration (e.g., a swivel office chair) can produce geometrically inconsistent correspondences.
Similar failures arise when the retrieved CAD differs substantially in shape from the observed object, or when deformable or weakly textured parts (e.g., cushioning) introduce geometric discrepancies.

\section{Conclusion}
\label{sec_conclusions}

We introduced SUFLECA, a weakly supervised zero-shot method for CAD-to-image alignment. Its two core contributions, unified NOC-aligned feature learning across diverse real and synthetic data and mutual \(k\)-nearest-neighbor matching with geometric consensus filtering, jointly address the synthetic-to-real domain gap and spatial inconsistencies of prior approaches. Together, they improve robustness to appearance variation while suppressing semantically plausible but geometrically inconsistent correspondences. SUFLECA uses compact yet discriminative features to enable sub-second alignment without direct pose supervision from the target benchmark or iterative refinement, providing an efficient approach for CAD-based scene understanding. Extending it to broader object categories and improving robustness to articulation, deformation, and inexact CAD retrieval remain important for open-world deployment.


\clearpage
\clearpage

\thispagestyle{empty}
{\centering\bfseries\Large\title Supplementary Material} 
\section*{Overview of content}
This supplementary material provides the following:
\begin{itemize}
\item \textbf{~\Cref{sec:training}} -- Additional details on the SUFLECA feature model training setup.
\item \textbf{~\Cref{sec:alignment}} -- Configuration for the alignment pipeline.
\item \textbf{~\Cref{sec:competing}} -- Implementation details for competing open-source methods.
\item \textbf{~\Cref{sec:metrics}} -- Descriptions of the metrics used for evaluation on the CO3D dataset.
\item \textbf{~\Cref{sec:replica}} -- Additional details on the ReplicaCAD dataset and setup.
\item \textbf{~\Cref{sec:wild}} -- An in-the-wild CAD-to-image alignment pipeline based on SUFLECA.
\item \textbf{~\Cref{sec:sam3d}} -- A comparison with model-free methods for novel object pose estimation.
\item \textbf{~\Cref{sec:tradeoff}} -- Analysis of accuracy-speed tradeoff.
\item \textbf{~\Cref{sec:complementary}} -- Complementarity of performance improvements from our contributions.
\item \textbf{~\Cref{sec:detailed}} -- A detailed analysis of the ScanNet25k validation results reported in the main paper, including qualitative evaluations.
\end{itemize}

\section{NOC-aligned feature learning}
\label{sec:training}
\subsection{Parameter Counts}

SUFLECA, which uses DUNE-B~\cite{sariyildiz2025dune} as the frozen perception encoder, comprises 101.5M parameters, of which 15.2M are trainable.
SUFLECA-S, which uses the smaller DUNE-S backbone, comprises 28.6M parameters, with 6.6M trainable.

\subsection{Dataset statistics}
For ARKitScenes~\cite{dehghan2021arkitscenes}, REAL275~\cite{NOCS}, and Hypersim~\cite{hypersim}, we use the frames and annotations provided by OmniNOCS~\cite{krishnan2024omninocs}.
Objectron~\cite{objectron2021} is also provided through OmniNOCS, but we retain only a subset of approximately 7k frames rather than the full 132k, owing to data availability constraints and to avoid disproportionately biasing the training set toward Objectron categories.
For ScanNet~\cite{dai2017scannet} and ScanNet++~\cite{yeshwanth2023scannet++}, we sample every 30th frame and use the annotations provided by Scannotate~\cite{scannotatepp}.
For RealEstate10K~\cite{realestate} videos, we sample every 22nd frame and use the corresponding annotations from CAD-Estate~\cite{maninis2023cad}.
Owing to the slow camera motion in these videos, we additionally exclude frames whose camera displacement relative to the previously retained frame is below both $15\mathrm{cm}$ and $15^\circ$.
For Pix3D~\cite{pix3d}, ObjectNet3D~\cite{xiang2016objectnet3d}, and Pascal3D+~\cite{everingham2010pascal}, we process all available frames from indoor object categories, excluding outdoor categories such as \textit{car}, \textit{airplane}, and \textit{boat}, as well as the categories assigned to the \textit{unseen} group in our CO3D evaluation and their synonymous labels.
For 3D-FRONT~\cite{3dfront}, we use BlenderProc~\cite{blenderproc} to render 10 images from randomly sampled camera poses for each valid layout file.
For ShapeNet~\cite{chang2015shapenet}, we render random views of objects belonging to the 35 synsets that constitute the Scan2CAD~\cite{avetisyan2019scan2cad} CAD library.
Across all datasets, we remove frames in which annotated regions cover less than $2.5\%$ of the image pixels to improve training efficiency.
For real-world datasets with CAD annotations, namely ScanNet, ScanNet++, RealEstate10K, Pix3D, ObjectNet3D, and Pascal3D+, we render normalized object coordinates from the annotated CAD model using the corresponding camera pose and compare them against NOC annotations derived from the object pose and point cloud following~\cite{krishnan2024omninocs}.
We discard frames for which the mean Euclidean NOC error exceeds $0.1$.
Further, for each valid frame from these real-world datasets, we additionally generate a synthetic counterpart, as described in the main paper, to augment the training data and encourage the model to learn a unified feature representation across real and synthetic imagery.

For validation, we use the \textit{val} split of ARKitScenes~\cite{dehghan2021arkitscenes}, the \textit{test} split of REAL275, all valid frames from the ScanNet~\cite{dai2017scannet} \textit{validation} sequences, and frames from $10\%$ of the layouts provided by 3D-FRONT.
The complete dataset contains 674,851 frames, of which 634,754 are used for training and 40,097 for validation.
For the ScanNet25k~\cite{gumeli2022roca} evaluation, we train separate SUFLECA variants that exclude ScanNet frames during training.
After this exclusion, the dataset contains 557,567 frames, including 14,779 validation frames.

\subsection{Training details}
We train all SUFLECA variants for 50 epochs, with each epoch comprising 131,072 images.
We set the NOC loss weight to \(\lambda=0.33\) and optimize using AdamW~\cite{adamW} with a batch size of 16.
The learning rate is updated at every optimization step, with a linear warm-up from $1.0 \times 10^{-6}$ to $1.1 \times 10^{-4}$ over the first epoch, followed by cosine decay to $1.0 \times 10^{-6}$ across the remaining epochs.
We apply a weight decay of $3.0 \times 10^{-5}$ to all trainable parameters except biases and normalization parameters.

\section{Alignment configuration details}
\label{sec:alignment}
For alignment, we maintain $n=6$ fixed, pre-rendered views for each CAD model.
Before correspondence estimation, both the source and target point clouds are subsampled to at most $N=1024$ points.
We set the number of neighbors in the mutual $k$-nearest-neighbor matching procedure to $k=12$ and retain at most $M=256$ correspondences.
For geometric filtering, we use a relative inlier threshold of $\beta=0.25$ and a relative gating threshold of $\delta=0.40$ on the clique-agreement score.
The isotropic log-scale is the mode of a $60$-bin histogram over $\{\psi_{ij}\}$, with the bin count trading scale resolution against the number of pairs per bin.
As maximum clique is NP-hard and the density of $\mathbf{G}$ grows with $\beta$, we solve it following ~\cite{clique} under a budget of $20{,}000$ expansions, falling back to the best clique found so far, which remains a valid seed.

All alignment hyperparameters in this section are selected using a held-out set of 858 RealEstate10K~\cite{realestate} images. 
The selected parameters are then kept fixed across all evaluation benchmarks. 
All experiments are conducted on a workstation equipped with an AMD Ryzen 9 9950X CPU and an NVIDIA RTX 5090 GPU.

\subsection{Configuration selection and sensitivity analysis}
We tune four important parameters: the number of rendered views $n$, the number of neighbors $k$ used for geometric consistency filtering, the relative inlier threshold $\beta$, and the relative gating threshold $\delta$ applied before and after maximal-clique selection.
For fairness, we select these values on a separate validation set derived from RealEstate10K~\cite{realestate}.
The set contains 858 frames across nine categories (\textit{table}, \textit{sofa}, \textit{bed}, \textit{bathtub}, \textit{cabinet}, \textit{chair}, \textit{bookcase}, \textit{lamp}, and \textit{stove}) and 561 unique CAD models, with each category represented by 61--109 frames.
For ground-truth CAD annotations, we use CAD-Estate~\cite{maninis2023cad} and estimate approximate metric scale using MetricAnything~\cite{metricanything2026} over multiple frames.
We follow the evaluation thresholds of~\cite{gumeli2022roca}, considering an alignment correct if its translation, rotation, and scale errors are within the standard $20,\mathrm{cm}/20^\circ/20\%$ thresholds.
We use category-averaged alignment accuracy, referred to simply as \textit{Accuracy} in this section, as the primary criterion for parameter selection.
We additionally report the \textit{Success rate}, defined as the percentage of frames for which RANSAC produces a valid alignment within $300$k iterations.
Where relevant, we also consider the computational cost associated with each parameter.
All parameter sensitivity analyses are conducted using only the base SUFLECA variant.

\subsection{Render views}
Every CAD model in the retrieval pool is rendered from a fixed set of
$n$ viewpoints, shared across all SUFLECA variants.
Working in the model's canonical object frame, with the gravity axis along $+\mathbf{z}$, a viewing direction is parameterized by an azimuth $\theta$ and a polar angle $\phi$,
\begin{equation}
  \mathbf{d}(\theta,\phi) =
  \big(\sin\phi\cos\theta,\;\sin\phi\sin\theta,\;\cos\phi\big)^{\!\top}.
\end{equation}
The camera is placed at $\mathbf{c}_i = \mathbf{o} + r\mathbf{d}(\theta_i,\phi_i)$, looking toward the object center $\mathbf{o}$, where $r = 1.75\ell + 0.04$ and $\ell$ denotes the largest extent of the CAD model. 
We use azimuth angles
$\theta_i \in \{18^\circ,82^\circ,146^\circ,211^\circ,287^\circ,332^\circ\}$,
wih corresponding polar angles
$\phi_i \in \{58^\circ,63^\circ,56^\circ,61^\circ,54^\circ,60^\circ\}$.
We use the SAPIEN~\cite{Xiang_2020_SAPIEN} simulator to render each view
with a randomized background color.

\begin{figure}[t]
    \centering
    \includegraphics[width=\linewidth]{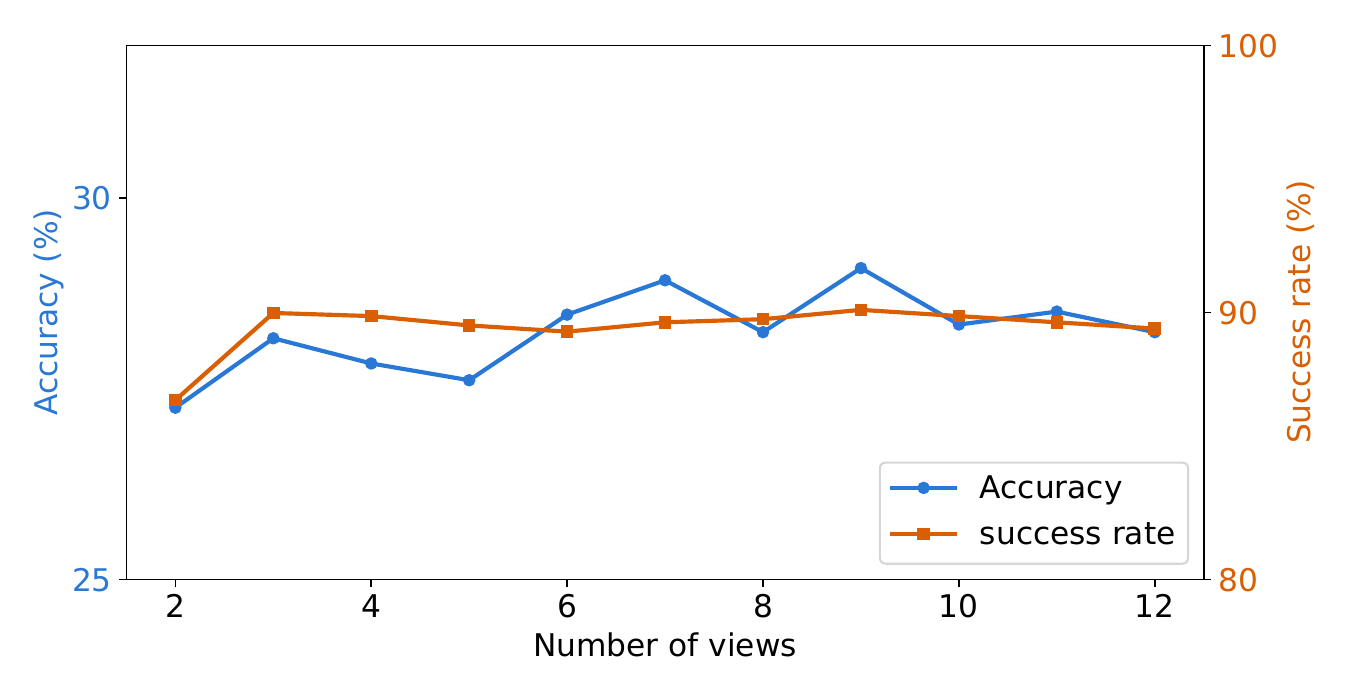}
    \caption{Sensitivity of alignment accuracy  and success rate to the number of pre-rendered CAD views $n$ on the DiffCAD~\cite{diffcad} split of ScanNet25k~\cite{gumeli2022roca}. Each view adds approximately $2.4\mathrm{ms}$ of alignment runtime and $6.43\mathrm{MB}$ disk space per CAD model in the offline precomputed library.}
    \label{fig:views}
\end{figure}

In \Cref{fig:views}, we evaluate the sensitivity of alignment accuracy and success rate to the number of rendered views.
Performance remains relatively stable across a broad range of view counts, indicating that, unlike prior works~\cite{wu2024diorama}, SUFLECA does not require dense viewpoint sampling.
The best performance is obtained with approximately $6$--$10$ views.
We therefore use $n=6$, avoiding the additional storage and runtime overhead of denser view sampling despite its relatively small cost.

\subsection{Number of neighbors}

\begin{figure}[t]
    \centering
    \includegraphics[width=\linewidth]{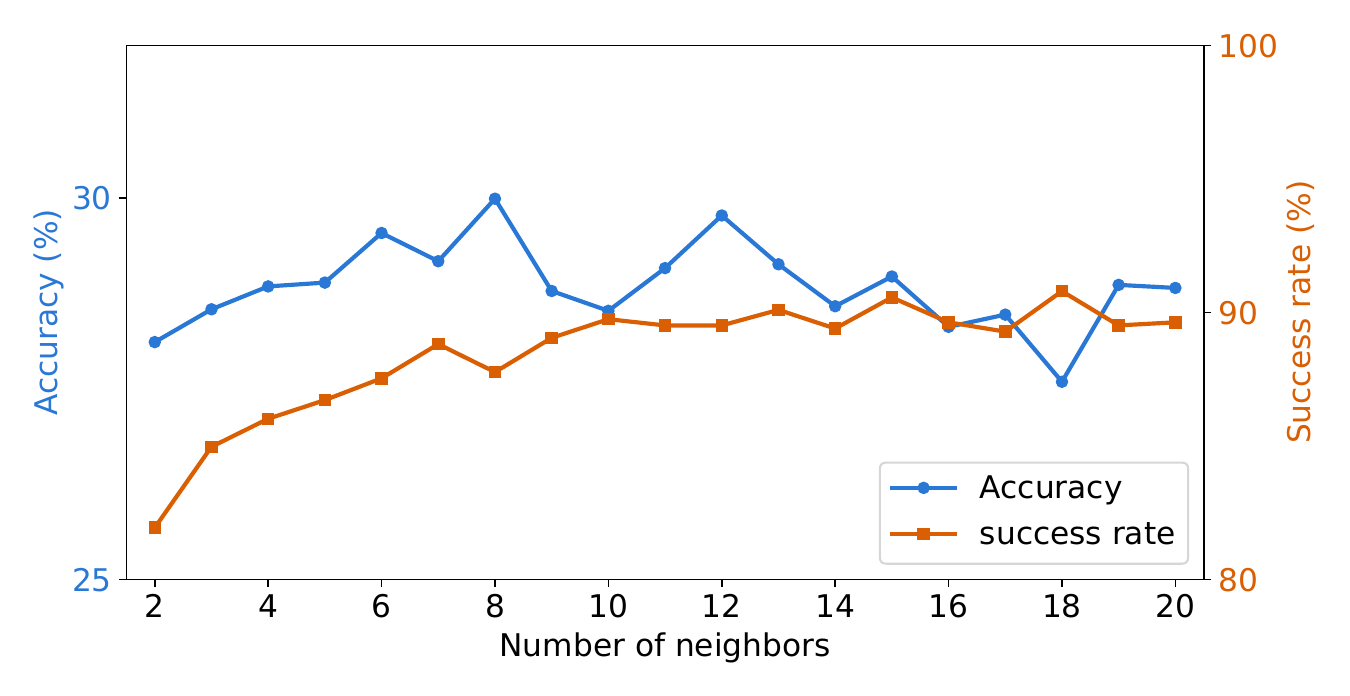}
    \caption{Sensitivity of the number of neighbors in our mutual $k$-nearest-neighbor algorithm.}
    \label{fig:neighbors}
\end{figure}
In \Cref{fig:neighbors}, we evaluate the sensitivity of alignment accuracy and success rate to the number of neighbors $k$ used in mutual $k$-nearest-neighbor matching.
Small values ($k<10$) achieve comparable accuracy but substantially lower success rates, as the mutual constraint retains too few correspondences, and subsequent filtering can leave insufficient matches for RANSAC to estimate a valid registration.
We therefore select $k=12$, which provides a good balance of accuracy and success rate.
The number of neighbors has a negligible effect on runtime and is therefore not reported.

\subsection{Relative thresholds}

\begin{figure}[t]
    \centering
    \includegraphics[width=\linewidth]{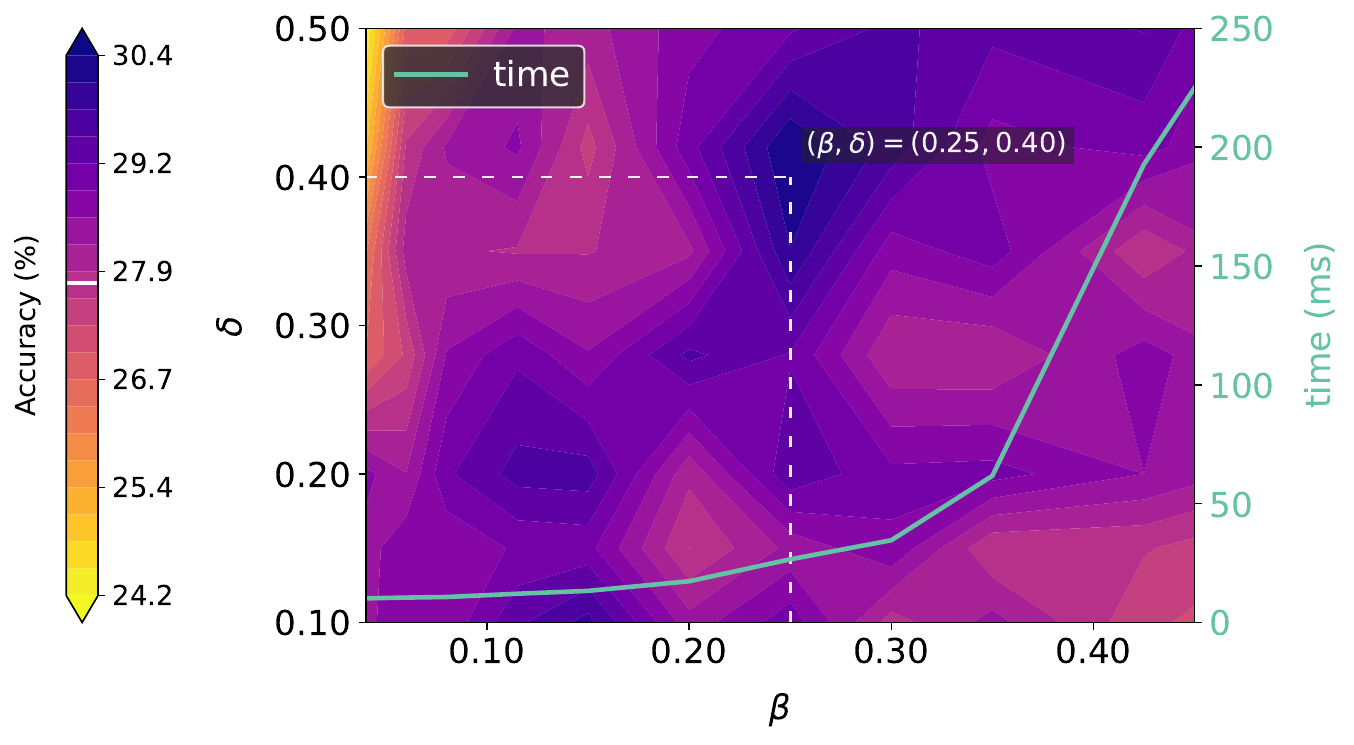}
    \caption{Contour plot of alignment accuracy for varying relative thresholds $\beta$ and $\delta$.
    The overlaid green curve shows runtime in $\mathrm{ms}$, which increases monotonically with $\beta$.
    The horizontal white marker on the accuracy color bar indicates the $27.8\%$ accuracy obtained without any geometric consistency filtering.}
    \label{fig:beta_rel}
\end{figure}
In \Cref{fig:beta_rel}, we evaluate the sensitivity of alignment accuracy and runtime to the relative thresholds $\beta$ and $\delta$.
Compared with the $27.8\%$ accuracy obtained without geometric consistency filtering, many threshold combinations improve performance, indicating that the method is not highly sensitive to a narrow operating point.
We select $\beta=0.25$ and $\delta=0.40$, which achieve the highest evaluated accuracy within the largest high-performing region of the contour plot, indicated by the darker blue area.
We use this criterion to avoid selecting an isolated parameter combination that may be overfit to the chosen RealEstate10K frames.
As $\beta$ increases, the graph $\mathbf{G}$ becomes denser, increasing the cost of maximal-clique search and the likelihood of reaching its iteration budget.
However, at the selected operating point, this cost remains modest at well below $50\,\mathrm{ms}$.

\section{Details on competing methods}
\label{sec:competing}
\subsection{DINOv3-L/DUNE-B}
For the unsupervised feature baselines~\cite{simeoni2025dinov3,sariyildiz2025dune}, we establish correspondences using the nearest-neighbor matching strategy employed by ZeroCAD~\cite{Arsomngern2025ZeroCAD} and ROCA~\cite{gumeli2022roca}, assigning each source point to its nearest target point in feature space.
To ensure a fair comparison and bound the runtime, we uniformly subsample both the source and target point sets to at most 8,192 points before correspondence estimation.
We discard correspondences with cosine similarity below $0.5$ and retain at most 512 correspondences using farthest-point sampling in 3D.
In our experiments, this sampling strategy improves accuracy while reducing run-to-run variance.

\subsection{Diorama}
For Diorama~\cite{wu2024diorama}, we follow the official implementation while using 180 rendered templates per retrieved CAD model for coarse pose estimation, with GigaPose~\cite{nguyen2024gigapose} providing the object scale prior.
Our reproduced ScanNet25k results are higher than those reported in the original paper, potentially due to the increased number of render templates and improvements in object segmentation and depth estimation.
We report these improved reproduced results throughout our evaluation. 

\subsection{FoundationPose}
FoundationPose~\cite{wen2024foundationpose} estimates only the 6D object pose and does not recover the scale.
We therefore report the 9D extension presented by ZeroCAD~\cite{Arsomngern2025ZeroCAD}, which augments FoundationPose with supervised scale estimates from ROCA~\cite{gumeli2022roca}.

\subsection{SDFit}
SDFit~\cite{antic2025sdfit} is a recent method for category-level pose and shape fitting.
However, using the official implementation, we were unable to reproduce competitive results under the ScanNet25k~\cite{gumeli2022roca} evaluation protocol.
Furthermore, SDFit requires between three and five minutes of optimization per object and is highly sensitive to the accuracy of the initial pose estimate, making it impractical for the large-scale evaluations considered in this work.
Therefore, we do not include it in any of the experiments.



\subsection{Runtime-only proxy of ZeroCAD}
Since reproducing the ZeroCAD~\cite{Arsomngern2025ZeroCAD} feature model would require regenerating its training data and retraining the adapter, we instead report the published ZeroCAD results on the common ScanNet25k~\cite{gumeli2022roca} and DiffCAD~\cite{diffcad} splits.
For runtime evaluation, we construct a proxy implementation that provides an approximate estimate of ZeroCAD's runtime.
Specifically, we replace its adapted DINOv2~\cite{oquab2023dinov2} features with concatenated DINOv2 features, thereby excluding the computational overhead of the ZeroCAD DINOv2 adapter and yielding an optimistic runtime estimate for this stage.
For the refinement stage, we use a Gaussian-splatting-based renderer to approximate the render-and-compare optimization procedure used by ZeroCAD.
Specifically, the CAD surface is subsampled using farthest-point sampling and represented with a differentiable soft rasterizer, where truncated isotropic Gaussians are accumulated into a silhouette buffer and a normalized soft $z$-buffer.
This produces differentiable rendered masks and depth maps with respect to all nine alignment parameters.
The optimization objective is a weighted combination of three terms: an $\ell_{1}$ silhouette loss against the instance mask, an $\ell_{1}$ depth loss over jointly valid pixels to constrain translation along the optical axis, and a unidirectional Chamfer loss from the back-projected scene point cloud to the transformed CAD surface.
We optimize this objective using Adam for 500 iterations, as this was empirically found on a small held-out subset of ScanNet25k to be sufficient for reaching a stable loss basin.
This refinement stage takes approximately 2--3 seconds per object.
A direct runtime comparison with the original ZeroCAD is not possible, as the paper does not report timing analyses, and its implementation is not publicly available.

\section{Metrics for CO3D evaluation}
\label{sec:metrics}
Because CO3D~\cite{reizenstein21co3d} does not provide a shared canonical
object orientation across instances, the standard rotation, translation, and
scale errors used for ScanNet25k are not directly applicable.
We therefore adopt the geometry-based evaluation protocol of
SAM3D~\cite{sam3dteam2025sam3d3dfyimages}, which compares predicted and
ground-truth object geometries in the CO3D world coordinate system without
requiring a common canonical object frame.

\subsection{3D-IoU}
3D-IoU is defined as the volumetric intersection-over-union of the
axis-aligned bounding boxes (AABBs) enclosing the predicted and ground-truth
object geometries in CO3D world coordinates.
This metric captures differences in object position, scale, and spatial
extent without requiring correspondence between the canonical coordinate
systems of the two objects.

\subsection{ICP-Rot}
To evaluate orientation without assuming a shared canonical frame, we rigidly
register the predicted object geometry to the ground-truth geometry using
Iterative Closest Point (ICP).
ICP-Rot is the angular magnitude of the rotational correction estimated by
ICP and is reported in degrees.
Lower values indicate that less rotational correction is required to align the
predicted geometry with the ground-truth geometry.

\subsection{ADD-S}
ADD-S measures the symmetric average closest-point distance between the
predicted and ground-truth object surfaces, normalized by the diameter of the
ground-truth object.
The symmetric formulation averages closest-point distances in both directions.
We sample $5{,}000$ points from each object surface using farthest-point
sampling to compute the metric.
Lower values indicate a better geometric agreement between the predicted and
ground-truth object surfaces.

\subsection{ADD-S@0.1}
ADD-S@0.1 is the percentage of predictions whose ADD-S error is below $0.1$.
Higher values indicate a larger fraction of accurately aligned instances.

\section{Additional details on ReplicaCAD}
\label{sec:replica}
As described in the main paper, we use ReplicaCAD~\cite{szot2021habitat}
as a controlled synthetic setting to analyze the effect of real and synthetic
data on SUFLECA's learned feature representation. ReplicaCAD is not used for
model selection or as an additional comparison benchmark; rather, it
complements the real-world ScanNet25k evaluation in the training-data
composition analysis.
We render 3,457 ReplicaCAD frames across 90 scene configurations, covering
1,087 object instances from twelve categories: table, chair, plant, cabinet,
display, box, sofa, stool, trash bin, vase, basket, and lamp.
Each category is represented by 112--685 frames.
Since ReplicaCAD contains relatively few instances of some categories, we
sample additional viewpoints at three camera elevations for the sparsest
categories so that each category contains more than 100 frames.
CAD retrieval is performed over all 92 ReplicaCAD assets using our zero-shot strategy, described later in ~\Cref{sec:zero_retr}.

For category- and instance-averaged alignment accuracy, we follow the same
evaluation criterion as ScanNet25k~\cite{gumeli2022roca}, considering an
alignment correct when its translation, rotation, and scale errors are within
20\,cm, $20^\circ$, and 20\%, respectively.
We additionally report 3D-IoU, ICP-Rot, and ADD-S to provide complementary
geometry-based measures of alignment quality.

ReplicaCAD does not provide object-symmetry annotations required by the
rotation component of the threshold-based alignment metric.
We therefore derive rotational symmetries geometrically by rotating each mesh
about its upright axis and measuring its symmetric Chamfer residual against
the original mesh.
The resulting symmetry estimates are used only for the threshold-based pose
evaluation. The additional geometry-based metrics provide complementary
evaluation that is not dependent on these derived annotations.

\section{Additional analysis with a fully zero-shot pipeline}
\label{sec:wild}

The main evaluation follows prior work and uses fixed upstream components,
including object detections and CAD retrievals provided by the supervised
ROCA method~\cite{gumeli2022roca}, to isolate CAD alignment performance and
ensure direct comparability on ScanNet25k~\cite{gumeli2022roca}.
As noted in the main paper, we additionally evaluate SUFLECA within a fully
zero-shot pipeline in this supplementary analysis.
Specifically, we replace the supervised detection and CAD retrieval components
with zero-shot alternatives while retaining the SUFLECA alignment stage.
None of the components in this pipeline are trained or fine-tuned on
ScanNet25k (for depth estimation, we use the off-the-shelf MetricAnything~\cite{metricanything2026} model without ScanNet25k fine-tuning).
This analysis uses the same ScanNet25k evaluation setting as the main paper
and studies how upstream retrieval quality affects SUFLECA within a fully
zero-shot pipeline.
We first describe the zero-shot CAD retrieval strategy in \Cref{sec:zero_retr}. 
We then evaluate the resulting pipeline on the complete ScanNet25k validation split in \Cref{sec:inexact} to analyze SUFLECA's robustness to the substantially
noisier CAD retrievals obtained without dataset-specific supervision.

\subsection{Zero-shot CAD retrieval}
\label{sec:zero_retr}

We adapt the open-set CAD retrieval pipeline from
OSCAR~\cite{pulli2026oscar} into a purely visual pipeline.
First, we define an object vocabulary of common category names mapped to ShapeNet~\cite{chang2015shapenet} synsets by prompting an LLM (GPT-5.5) with "\textit{Given the ShapeNet synset taxonomy, identify common object categories that may appear in indoor scenes. For each category, provide a concise everyday English name and its corresponding ShapeNet synset. Do not introduce categories that are not represented in ShapeNet.}" The resulting vocabulary is manually checked to verify the category names and ShapeNet mappings, and is kept fixed across all experiments.
GroundingDINO~\cite{liu2024grounding} uses this vocabulary to detect candidate
object bounding boxes, which are subsequently segmented using
SAM2~\cite{ravi2024sam2}.
The predicted object category is then used to restrict the CAD search space to
the corresponding ShapeNet synset.

Retrieval proceeds in two stages.
We use DINOv3-L~\cite{simeoni2025dinov3} features and first obtain a coarse ranking.
For each CAD model, 48 pre-rendered and pre-featurized template views are
compared with the masked input image using generalized mean embeddings,
requiring only a single-vector comparison per template view.
This stage produces the top 100 CAD-view hypotheses, which are then reranked using patch-level feature correspondence matching with the same features computed during the coarse stage.
Specifically, we compute the mean nearest-neighbor similarity between
patch-level masked image features and each candidate render and rank CAD
models by the aggregated similarity across their rendered views.

We evaluate several feature representations for this template-based retrieval
pipeline, including TIPSv2-SO~\cite{tips_v2_paper},
DUNE-B~\cite{sariyildiz2025dune},
DINOv3-L~\cite{simeoni2025dinov3}, and our SUFLECA embeddings.
We additionally compare with OpenShape~\cite{liu2023openshape}, which retrieves
CAD models by matching CLIP~\cite{clip} image embeddings to precomputed 3D mesh
embeddings.
Based on the results in \Cref{tab:retrieval}, we select DINOv3-L as the
retrieval backbone because it achieves substantially higher top-1 and top-5
retrieval accuracy than the alternative representations, despite having a
slightly higher Chamfer Distance (CD) than SUFLECA and TIPSv2-SO.
Compared with supervised CAD retrieval from ROCA~\cite{gumeli2022roca},
zero-shot retrieval remains substantially less accurate, highlighting CAD
retrieval as a major bottleneck for in-the-wild CAD-based scene mapping.

This coarse-to-fine strategy enables efficient zero-shot retrieval at up to 20 objects per second on our hardware.
We use this retrieval strategy both to evaluate robustness to inexact CAD retrieval in \Cref{sec:inexact} and to retrieve candidate CADs from the SAM3D-generated CAD pool for the CO3D~\cite{reizenstein21co3d} evaluation
reported in the main paper.

\subsection{Robustness to inexact CAD retrieval}
\label{sec:inexact}

We further analyze how alignment accuracy depends on the quality of the
retrieved CAD model.
As shown in \Cref{tab:inexact}, replacing ground-truth Scan2CAD
annotations~\cite{avetisyan2019scan2cad} with zero-shot detection and retrieval
using GroundingDINO~\cite{liu2024grounding} and
OSCAR~\cite{pulli2026oscar} reduces category- and instance-averaged alignment
accuracy by $18.2$ and $20.1$ percentage points, respectively.
This degradation is accompanied by a particularly low instance retrieval
accuracy of 5.5\%, reflecting the difficulty of retrieving the correct CAD
model under heavy occlusion, appearance differences, and annotation bias.
These results suggest that zero-shot CAD retrieval from real-world
observations remains an open problem and a major bottleneck for a fully
zero-shot CAD fitting pipeline.
Nevertheless, as we move from the ground-truth oracle to supervised
ROCA~\cite{gumeli2022roca} and then to fully zero-shot retrieval, the decrease
in instance retrieval accuracy is substantially larger than the corresponding
decrease in alignment accuracy.
This suggests that SUFLECA retains moderate robustness to CAD inexactness,
although differences in object detection also contribute to the observed
performance gap.

\begin{table}[t]
    \centering
    \footnotesize
    \begin{tabular}{
    >{\raggedright\arraybackslash}p{2.2cm}|
    >{\centering\arraybackslash}p{0.75cm}
    >{\centering\arraybackslash}p{0.75cm}
    >{\centering\arraybackslash}p{0.75cm}|
    >{\centering\arraybackslash}p{0.75cm}}
        \toprule
        \multirow{2}{*}{Method}
        & \multicolumn{3}{c|}{Retrieval acc. (\%) $\uparrow$}
        & CD $\downarrow$ \\
        & top-1 & top-5 & top-10 & ($\times 100$) \\
        \midrule
        \textcolor{gray}{ROCA~\cite{gumeli2022roca}}
        & \textcolor{gray}{76.02}
        & \textcolor{gray}{--}
        & \textcolor{gray}{--}
        & \textcolor{gray}{1.68} \\
        \midrule
        OpenShape~\cite{liu2023openshape}
        & 4.02 & 12.84 & 18.99 & 6.28 \\
        DUNE-B~\cite{sariyildiz2025dune}
        & 3.53 & 11.98 & 15.63 & 6.22 \\
        DINOv3-L~\cite{simeoni2025dinov3}
        & \cellcolor{best!60}{\textbf{8.41}}
        & \cellcolor{best!60}{\textbf{21.49}}
        & \cellcolor{best!60}{\textbf{29.45}}
        & 5.81 \\
        TIPSv2-SO~\cite{tips_v2_paper}
        & \cellcolor{secondbest}{6.81}
        & \cellcolor{secondbest}{20.22}
        & \cellcolor{secondbest}{27.44}
        & \cellcolor{secondbest}{5.77} \\
        SUFLECA (\textit{Ours})
        & 4.06 & 15.50 & 22.07
        & \cellcolor{best!60}{\textbf{5.59}} \\
        \bottomrule
    \end{tabular}
    \caption{
    Zero-shot CAD retrieval performance of different methods and feature
    backbones on the DiffCAD~\cite{diffcad} split of
    ScanNet25k~\cite{gumeli2022roca}, using
    Scan2CAD~\cite{avetisyan2019scan2cad} annotations as ground truth.
    CD denotes the median Chamfer Distance between the retrieved and
    ground-truth CAD models in normalized ShapeNet~\cite{chang2015shapenet}
    coordinates, scaled by 100.
    ROCA~\cite{gumeli2022roca}, which is supervised with Scan2CAD annotations,
    is included as a reference.
    }
    \label{tab:retrieval}
\end{table}

\begin{table}[t]
    \footnotesize
    \centering
    \resizebox{\linewidth}{!}{%
    \begin{tabular}{
    >{\arraybackslash}p{3.0cm}|
    >{\centering\arraybackslash}p{1.0cm}
    >{\centering\arraybackslash}p{1.0cm}|
    >{\centering\arraybackslash}p{0.6cm}
    >{\centering\arraybackslash}p{0.6cm}}
        \toprule
        \multirow{2}{*}{Detection and retrieval}
        & \multirow{2}{*}{Det. acc.}
        & \multirow{2}{*}{Retr. acc.}
        & \multicolumn{2}{c}{Alignment acc.} \\
        &&& cat. & inst. \\
        \midrule
        Scan2CAD (GT)~\cite{avetisyan2019scan2cad}
        & 100.0 & 100.0 & 40.2 & 51.6 \\
        ROCA~\cite{gumeli2022roca}
        & 48.9 & 65.1 & 32.8 & 42.6 \\
        Zero-shot~\cite{liu2024grounding,pulli2026oscar}
        & 68.3 & 5.5 & 22.0 & 31.5 \\
        \bottomrule
    \end{tabular}}
    \caption{
    Effect of CAD retrieval quality on alignment accuracy on the complete
    ScanNet25k~\cite{gumeli2022roca} validation set.
    ROCA is supervised on ScanNet25k, whereas the zero-shot pipeline uses
    GroundingDINO~\cite{liu2024grounding} with our zero-shot CAD retrieval adapted from OSCAR~\cite{pulli2026oscar}.
    All methods use SAM2~\cite{ravi2024sam2} for instance segmentation.
    Det. acc. denotes detection accuracy while Retr. acc. denotes retrieval accuracy. All reported values are in \%.
    }
    \label{tab:inexact}
\end{table}

\section{Additional comparison with model-free novel object pose estimation methods}
\label{sec:sam3d}

\begin{table}[t]
    \centering
    \footnotesize
    \resizebox{\linewidth}{!}{%
    \begin{tabular}{
    >{\raggedright\arraybackslash}p{3.8cm}|
    >{\centering\arraybackslash}p{0.75cm}
    >{\centering\arraybackslash}p{0.75cm}|
    >{\centering\arraybackslash}p{1.0cm}
    >{\centering\arraybackslash}p{1.0cm}}
        \toprule
        \multirow{2}{*}{Method} & \multicolumn{2}{c|}{Alignment acc.} & VRAM & Per-inst.\\
        & cat. ↑ & inst. ↑ & (MB) ↓ &(s) ↓ \\
        \midrule
        SAM3D$^\dagger$~\cite{sam3dteam2025sam3d3dfyimages} & 11.5 & 16.9 & \cellcolor{secondbest}18627 & \cellcolor{secondbest}6.72 \\
        SAM3D$^\dagger$ (w/ refine)~\cite{sam3dteam2025sam3d3dfyimages} & \cellcolor{secondbest}20.4 & \cellcolor{secondbest}29.8 & 18642 & 7.08 \\
        SAM3D-OA (w/ refine)~\cite{sam3dteam2025sam3d3dfyimages,wangorient} & 18.5 & 26.4  & 23398 & 14.27 \\
        SUFLECA (\textit{Ours}) & \cellcolor{best!60}\textbf{25.3} & \cellcolor{best!60}\textbf{34.8} & \cellcolor{best!60}\textbf{5016} &  \cellcolor{best!60}\textbf{0.59} \\
        
        \bottomrule
    \end{tabular}}
    \caption{Alignment accuracy and efficiency comparison between SUFLECA with zero-shot CAD retrieval and SAM3D~\cite{sam3dteam2025sam3d3dfyimages} on the DiffCAD~\cite{diffcad} split of ScanNet25k~\cite{gumeli2022roca}.
    Accuracy metrics are reported in \%. VRAM is peak allocated GPU memory, while Per-inst. is the time taken to generate/retrieve a mesh/CAD and align it. Methods marked with $^\dagger$ require ground-truth poses to calibrate inconsistent canonical orientations.}
    \label{tab:sam3d}
\end{table}

\begin{table*}[t]
    \footnotesize
    \centering
    \resizebox{\linewidth}{!}{%
    \begin{tabular}{
    >{\arraybackslash}p{3.0cm}|
    >{\centering\arraybackslash}p{1.25cm}
    >{\centering\arraybackslash}p{1.25cm}
    >{\centering\arraybackslash}p{1.25cm}
    >{\centering\arraybackslash}p{1.65cm}|
    >{\centering\arraybackslash}p{1.25cm}
    >{\centering\arraybackslash}p{1.25cm}
    >{\centering\arraybackslash}p{1.25cm}
    >{\centering\arraybackslash}p{1.65cm}}
        \toprule
        \multirow{2}{*}{Method} & \multicolumn{4}{c|}{Seen categories} & \multicolumn{4}{c}{Unseen categories} \\
        & 3D-IoU ↑ & ICP-Rot ↓ & ADD-S ↓ & ADD-S@0.1↑ & 3D-IoU ↑ & ICP-Rot ↓ & ADD-S ↓ & ADD-S@0.1↑ \\
        \midrule
        \textcolor{gray}{SAM3D (w/ refine)~\cite{sam3dteam2025sam3d3dfyimages}}$^\ddagger$ & \textcolor{gray}{60.17} & \textcolor{gray}{15.00} & \textcolor{gray}{4.78} & \textcolor{gray}{82.33} & \textcolor{gray}{71.95} & \textcolor{gray}{9.80} & \textcolor{gray}{3.87} & \textcolor{gray}{91.73} \\
        \textcolor{gray}{SUFLECA (\textit{Ours})}$^\ddagger$ & \textcolor{gray}{66.08} & \textcolor{gray}{8.83} & \textcolor{gray}{3.63} & \textcolor{gray}{85.50} & \textcolor{gray}{53.03} & \textcolor{gray}{12.26} & \textcolor{gray}{5.17} & \textcolor{gray}{82.50} \\
        SAM3D~\cite{sam3dteam2025sam3d3dfyimages} & 46.35 & 17.34 & 5.44 & 82.50 & \cellcolor{secondbest}45.77 & \cellcolor{best!60}\textbf{11.23} & \cellcolor{secondbest}5.96 & \cellcolor{secondbest}79.75 \\
        SAM3D (w/ refine)~\cite{sam3dteam2025sam3d3dfyimages} & \cellcolor{secondbest}48.23 & \cellcolor{secondbest}16.43 & \cellcolor{secondbest}4.84 & \cellcolor{secondbest}83.50 & 41.28 & 14.92 & 6.11 & \cellcolor{best!60}\textbf{81.75} \\
        SUFLECA (\textit{Ours}) & \cellcolor{best!60}\textbf{62.30} & \cellcolor{best!60}\textbf{9.52} & \cellcolor{best!60}\textbf{3.77} & \cellcolor{best!60}\textbf{86.00} & \cellcolor{best!60}\textbf{48.77} & \cellcolor{secondbest}14.53 & \cellcolor{best!60}\textbf{5.73} & 79.25 \\
        \bottomrule
    \end{tabular}}
    \caption{Comparison between SAM3D and SUFLECA on selected CO3D~\cite{reizenstein21co3d} sequences comprising 600 images.
    Results for additional methods and SUFLECA variants are reported in the main paper. All methods use artificially occluded masks retaining approximately 67\% of the original mask area, except greyed out methods marked with $^\ddagger$, which use the full masks. 3D-IoU and ADD-S@0.1 are reported in \%, ICP-Rot in degrees, and ADD-S is scaled by 100.}
    \label{tab:co3d_supp}
\end{table*}

Works such as Any6D~\cite{lee2025any6d} and SAM3D~\cite{sam3dteam2025sam3d3dfyimages} demonstrate strong zero-shot mesh-to-image alignment capabilities but differ from CAD alignment methods in several important respects. Any6D relies on InstantMesh~\cite{xu2024instantmesh} to reconstruct a complete mesh from a single image, which can be challenging under occlusion or incomplete observations. Although Any6D can also accept an input mesh, we found its alignment sensitive to the anisotropic scaling required for retrieved, potentially inexact CAD models. SAM3D, in contrast, jointly generates and localizes a mesh during inference and therefore does not directly address alignment of a retrieved CAD model. Both approaches also incur the computational cost of mesh generation. Finally, generated meshes do not necessarily share a consistent canonical orientation across instances of a category, requiring either dataset-level canonicalization or an external orientation estimator such as OrientAnythingV2~\cite{wangorient}, as used by~\cite{jesslen2026geometrymatters}.

For additional context, we compare SAM3D~\cite{sam3dteam2025sam3d3dfyimages} with SUFLECA on the DiffCAD~\cite{diffcad} split of ScanNet25k~\cite{gumeli2022roca}. We evaluate both the base SAM3D pipeline and its post-inference refinement stage, which incorporates depth and silhouette losses, following the official implementation. Because SAM3D-generated meshes do not have a consistent canonical orientation across instances, we consider two evaluation protocols: (i) assigning a dominant canonical orientation separately for each category using the full evaluation set, denoted SAM3D$^\dagger$, and (ii) estimating the canonical orientation independently for each instance using OrientAnythingV2, denoted SAM3D-OA. For SUFLECA, we include the runtime and peak GPU memory required for zero-shot CAD retrieval from~\Cref{sec:zero_retr}, since SAM3D jointly generates and aligns a mesh and therefore does not require a separate CAD-retrieval stage. All methods use the object detections and segmentation masks provided by ROCA~\cite{gumeli2022roca} and monocular metric depth from MetricAnything~\cite{metricanything2026}. We do not report Any6D in this comparison because, in our preliminary experiments, we were unable to obtain reliable alignments under the clutter, occlusion, and segmentation noise present in ScanNet~\cite{dai2017scannet}. As shown in~\Cref{tab:sam3d}, SUFLECA outperforms both SAM3D$^\dagger$, which uses ground-truth pose information to determine category-level canonical orientations, and SAM3D-OA, which provides a practical alternative without such calibration. SUFLECA also requires substantially less runtime and GPU memory, largely because it avoids mesh generation. This distinction is particularly relevant when a retrieved CAD model can be reused across multiple observations.

For settings in which a shared canonical orientation is not required, such as CO3D~\cite{reizenstein21co3d}, we additionally compare against SAM3D using the same occluded masks and machine-generated depth inputs as SUFLECA in \Cref{tab:co3d_supp}. On \textit{seen} categories, SUFLECA consistently outperforms SAM3D across the reported alignment metrics. On \textit{unseen} categories, the methods are more comparable: SUFLECA achieves higher 3D-IoU, whereas SAM3D performs better on orientation error and thresholded surface accuracy. This may reflect SAM3D's broader training distribution, while SUFLECA's weak supervision is concentrated more heavily on common indoor object categories. We also evaluate SAM3D with the original, unoccluded masks (the occluded masks used in our evaluation retain approximately $67\%$ of the original mask area). With more complete observations, SAM3D improves substantially on the \textit{unseen} categories, while SUFLECA exhibits a smaller difference between the complete- and occluded-mask settings. 

These results suggest that expanding SUFLECA's weak supervision to less common categories, potentially using generated 3D predictions from methods such as SAM3D as pseudo-labels, is a promising direction for future work.

\section{Accuracy--Speed Trade-off}
\label{sec:tradeoff}

\begin{figure}[t]
    \centering
    \includegraphics[width=\linewidth]{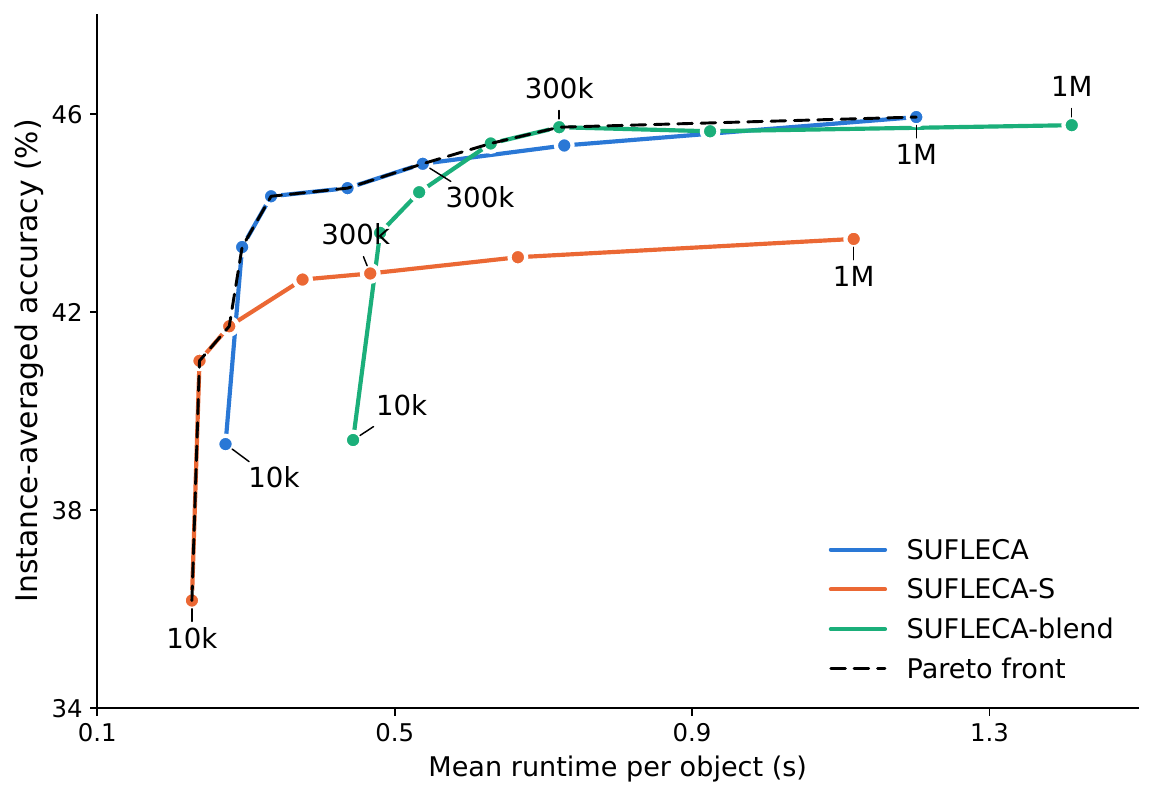}
    \caption{Accuracy--speed trade-off for SUFLECA variants under different maximum numbers of RANSAC iterations on the DiffCAD~\cite{diffcad} split of ScanNet25k~\cite{gumeli2022roca}.}
    \label{fig:ransac}
\end{figure}

As mentioned in the main paper, most of the alignment runtime is spent in RANSAC, yielding a direct trade-off between accuracy and efficiency.
We investigate this by evaluating all SUFLECA variants with different maximum numbers of RANSAC iterations on the DiffCAD~\cite{diffcad} split of ScanNet25k~\cite{gumeli2022roca}, as shown in \Cref{fig:ransac}.
SUFLECA provides the best accuracy across practical runtime budgets, while SUFLECA-S remains attractive under tighter resource constraints due to its lower GPU memory footprint and competitive accuracy at very low runtimes ($\sim0.25\,\mathrm{s}$/object).
SUFLECA-blend is designed for stronger generalization, as demonstrated by the CO3D evaluations in the main paper; however, this advantage is not fully reflected here because DiffCAD covers only six common object categories.
Nevertheless, all variants contribute to the Pareto front, enabling a choice between accuracy, runtime, memory efficiency, and generalization depending on deployment requirements.

\section{Complementarity of feature learning and correspondence estimation}
\label{sec:complementary}

To isolate the contributions of feature learning and correspondence estimation, we evaluate all four combinations of DUNE-B~\cite{sariyildiz2025dune} and SUFLECA features with nearest-neighbor (NN) matching~\cite{gumeli2022roca,Arsomngern2025ZeroCAD} and our correspondence estimator, while keeping the remainder of the alignment pipeline fixed.
The results on the ScanNet25k~\cite{gumeli2022roca} validation split in~\Cref{tab:component_complementarity} show substantial and largely complementary gains.
Replacing only the correspondence estimator improves category- and instance-averaged accuracy by $+9.0$ and $+13.0$ points, respectively, while replacing only the feature model yields gains of $+14.9$ and $+17.6$ points.
Combining both improves accuracy by $+21.4$ and $+26.0$ points, corresponding to approximately $90\%$ and $85\%$, respectively, of the improvement expected if the individual gains were fully additive.
This near-additive behavior suggests limited redundancy between the two contributions. The learned representation improves the quality of individual feature matches, while the correspondence estimator selects a geometrically consistent set of matches for alignment.

\begin{table}[t]
\centering
\resizebox{\linewidth}{!}{
    \begin{tabular}{cc|ccc}
    \toprule
    \textbf{Feature Model} & \textbf{Corr. Estimation} &
    \textbf{Cat. Acc.} $\uparrow$ & \textbf{Inst. Acc.} $\uparrow$ & \textbf{$\Delta$ Cat./Inst.} \\
    \midrule
    DUNE-B~\cite{sariyildiz2025dune} & NN~\cite{gumeli2022roca,Arsomngern2025ZeroCAD} & 11.4 & 16.6 & -- \\
    DUNE-B~\cite{sariyildiz2025dune} & \textit{Ours} & 20.4 & 29.6 & +9.0/+13.0\\
    \textit{Ours} & NN~\cite{gumeli2022roca,Arsomngern2025ZeroCAD} & 26.3 & 34.2 & +14.9/+17.6\\
    \textit{Ours} & \textit{Ours} & 32.8 & 42.6 & +21.4/+26.0 \\
    \bottomrule
    \end{tabular}}
    \caption{
        Ablation of the feature model and correspondence estimator on the ScanNet25k~\cite{gumeli2022roca} validation set.
        We report category- and instance-averaged alignment accuracy.
        The final column shows the absolute improvement in category/instance accuracy over the DUNE-B~\cite{sariyildiz2025dune} + nearest-neighbor (NN)~\cite{gumeli2022roca,Arsomngern2025ZeroCAD} baseline.
    }
    \label{tab:component_complementarity}
\end{table}

\section{Detailed analysis on ScanNet25k}
\label{sec:detailed}

\subsection{Error metrics breakdown}
\begin{table*}[t]
    \footnotesize
    \centering
    \resizebox{\textwidth}{!}{%
    \begin{tabular}{p{2.8cm}|
    >{\centering\arraybackslash}p{0.6cm}|
    >{\centering\arraybackslash}p{0.7cm}
    >{\centering\arraybackslash}p{0.7cm}
    >{\centering\arraybackslash}p{0.7cm}|
    >{\centering\arraybackslash}p{0.7cm}
    >{\centering\arraybackslash}p{0.7cm}
    >{\centering\arraybackslash}p{0.7cm}|
    >{\centering\arraybackslash}p{0.7cm}
    >{\centering\arraybackslash}p{0.7cm}
    >{\centering\arraybackslash}p{0.7cm}|
    >{\centering\arraybackslash}p{0.7cm}
    >{\centering\arraybackslash}p{0.7cm}
    >{\centering\arraybackslash}p{0.7cm}|
    >{\centering\arraybackslash}p{0.7cm}
    >{\centering\arraybackslash}p{0.7cm}
    >{\centering\arraybackslash}p{0.7cm}}
        \toprule
        \multirow{2}{*}{\textbf{Method}} & \multirow{2}{*}{\textbf{Sup.}} & \multicolumn{3}{c|}{\textbf{bathtub}} & \multicolumn{3}{c|}{\textbf{bed}} & \multicolumn{3}{c|}{\textbf{bin}} & \multicolumn{3}{c|}{\textbf{bkshlf}} & \multicolumn{3}{c}{\textbf{cabinet}}\\
        && \textbf{trans.} &\textbf{ rot. }& \textbf{scale }& \textbf{trans. }& \textbf{rot.} &\textbf{ scale} & \textbf{trans.} & \textbf{rot.} & \textbf{scale} &
        \textbf{trans.} & \textbf{rot.} & \textbf{scale} & \textbf{trans.} & \textbf{rot.} & \textbf{scale} \\
        \midrule
        ROCA~\cite{gumeli2022roca} & \cmark & 19.5 & \cellcolor{secondbest}13.9 & \cellcolor{best!60}\textbf{15.9} & 37.6 & 15.0 & \cellcolor{best!60}\textbf{11.0} & 16.6 & \cellcolor{best!60}\textbf{15.0} & \cellcolor{best!60}\textbf{15.3} & 39.2 & \cellcolor{secondbest}11.9 & 27.1 & 42.2 & 10.5 & \cellcolor{best!60}\textbf{20.5} \\
        SPARC~\cite{sparc} & \cmark & 18.8 & \cellcolor{best!60}\textbf{9.1} & \cellcolor{secondbest}18.4 & 23.2 & 13.0 & \cellcolor{secondbest}13.9 & 14.7 & 17.8 & \cellcolor{secondbest}16.0 & 35.6 & \cellcolor{best!60}\textbf{6.9} & 27.2 & 42.5 & \cellcolor{best!60}\textbf{5.9} & 22.4 \\
        DINOv3-L~\cite{simeoni2025dinov3} & -- & 26.0 & 50.1 & 34.2 & 62.5 & 69.8 & 47.4 & 13.3 & 63.9 & 42.3 & 44.2 & 65.5 & 55.7 & 52.3 & 70.4 & 54.6 \\
        DUNE-B~\cite{sariyildiz2025dune} & -- & 22.5 & 77.4 & 34.9 & 61.6 & 56.9 & 49.3 & 9.6 & 48.6 & 32.7 & 48.7 & 53.4 & 60.4 & 48.5 & 59.7 & 51.3 \\
        Diorama~\cite{wu2024diorama} & -- & 29.2 & 56.7 & 35.5 & 56.9 & 46.9 & 46.9 & 10.6 & 27.5 & 22.2 & 42.6 & 32.6 & 44.2 & 57.9 & 78.4 & 34.4 \\
        SUFLECA (\textit{Ours}) & * & \cellcolor{best!60}\textbf{9.5} & 15.1 & 18.6 & \cellcolor{secondbest}20.1 & \cellcolor{best!60}\textbf{9.0} & 18.0 & \cellcolor{secondbest}6.9 & 17.8 & 16.6 & \cellcolor{best!60}\textbf{30.0} & 14.8 & 26.2 & \cellcolor{secondbest}38.6 & 9.9 & 21.9 \\
        SUFLECA-S (\textit{Ours}) & * & \cellcolor{secondbest}10.1 & 15.8 & 19.6 & \cellcolor{best!60}\textbf{20.0} & \cellcolor{secondbest}9.1 & 19.2 & \cellcolor{best!60}\textbf{6.4} & \cellcolor{secondbest}15.9 & 17.6 & 32.3 & 14.0 & \cellcolor{best!60}\textbf{25.2} & \cellcolor{best!60}\textbf{37.6} & 9.3 & 22.6 \\
        SUFLECA-blend (\textit{Ours}) & * & 12.0 & 14.6 & 19.0 & 24.0 & 9.6 & 19.3 & 7.3 & 21.7 & 16.7 & \cellcolor{secondbest}32.1 & 14.3 & \cellcolor{secondbest}26.1 & 38.9 & \cellcolor{secondbest}9.2 & \cellcolor{secondbest}20.8 \\
        \midrule
        \midrule
        \multirow{2}{*}{\textbf{Method}} & \multirow{2}{*}{\textbf{Sup.}} & \multicolumn{3}{c|}{\textbf{chair}} & \multicolumn{3}{c|}{\textbf{display}} & \multicolumn{3}{c|}{\textbf{sofa}} & \multicolumn{3}{c|}{\textbf{table}} & \multicolumn{3}{c}{\textbf{average}}\\
        && \textbf{trans.} &\textbf{ rot. }& \textbf{scale }& \textbf{trans. }& \textbf{rot.} &\textbf{ scale} & \textbf{trans.} & \textbf{rot.} & \textbf{scale} &
        \textbf{trans.} & \textbf{rot.} & \textbf{scale} & \textbf{trans.} & \textbf{rot.} & \textbf{scale} \\
        \midrule
        ROCA~\cite{gumeli2022roca} & \cmark & 20.0 & 13.6 & \cellcolor{best!60}\textbf{9.6} & 17.4 & 17.4 & 21.7 & 33.6 & 11.1 & 20.9 & 35.3 & 12.3 & 18.7 & 29.0 & 13.4 & \cellcolor{best!60}\textbf{17.9} \\
        SPARC~\cite{sparc} & \cmark & 15.8 & 9.5 & \cellcolor{secondbest}10.0 & 14.8 & 15.5 & 26.4 & 22.7 & \cellcolor{secondbest}6.7 & 16.0 & 32.1 & 13.2 & \cellcolor{secondbest}18.6 & 24.5 & \cellcolor{best!60}\textbf{10.8} & 18.8 \\
        DINOv3-L~\cite{simeoni2025dinov3} & -- & 16.4 & 31.5 & 27.8 & 10.7 & 31.6 & 33.5 & 25.7 & 15.0 & 21.3 & 50.0 & 65.6 & 56.9 & 33.5 & 51.5 & 41.5 \\
        DUNE-B~\cite{sariyildiz2025dune} & -- & 17.1 & 31.4 & 26.2 & 12.7 & 45.3 & 28.7 & 34.6 & 16.0 & 27.4 & 49.8 & 54.0 & 52.2 & 33.9 & 49.2 & 40.3 \\
        Diorama~\cite{wu2024diorama} & -- & 30.3 & 54.0 & 24.0 & 15.9 & 52.7 & 34.9 & 48.6 & 48.2 & 42.5 & 50.9 & 47.2 & 37.3 & 38.1 & 49.3 & 35.8 \\
        SUFLECA (\textit{Ours}) & * & \cellcolor{best!60}\textbf{8.4} & \cellcolor{best!60}\textbf{8.7} & 11.4 & 8.9 & \cellcolor{best!60}\textbf{13.4} & \cellcolor{best!60}\textbf{20.3} & \cellcolor{secondbest}14.7 & 6.9 & \cellcolor{best!60}\textbf{11.3} & \cellcolor{secondbest}26.8 & \cellcolor{best!60}\textbf{10.5} & \cellcolor{best!60}\textbf{17.6} & \cellcolor{best!60}\textbf{18.2} & \cellcolor{secondbest}11.8 & \cellcolor{secondbest}18.0 \\
        SUFLECA-S (\textit{Ours}) & * & 9.0 & 9.4 & 11.9 & \cellcolor{secondbest}8.7 & 15.1 & \cellcolor{secondbest}21.4 & \cellcolor{best!60}\textbf{14.0} & \cellcolor{best!60}\textbf{6.5} & \cellcolor{secondbest}12.6 & 28.8 & 13.2 & 19.2 & \cellcolor{secondbest}18.6 & 12.0 & 18.8 \\
        SUFLECA-blend (\textit{Ours}) & * & \cellcolor{secondbest}8.9 & \cellcolor{secondbest}8.8 & 12.3 & \cellcolor{best!60}\textbf{8.4} & \cellcolor{secondbest}13.9 & 24.1 & 16.5 & 7.2 & 13.8 & \cellcolor{best!60}\textbf{26.1} & \cellcolor{secondbest}10.9 & 18.9 & 19.4 & 12.3 & 19.0 \\
        \bottomrule
    \end{tabular}}
    \caption{Detailed alignment error analysis on the ScanNet25k~\cite{gumeli2022roca} evaluation (lower is better for all metrics). \textit{trans.} denotes the translation error in $\mathrm{cm}$, \textit{rot.} the rotation error in $^\circ$, and \textit{scale} the relative scale error in \%. Results are reported as the median error for each category, together with the mean of the per-category medians.}
    \label{tab:breakdown}
\end{table*}

We provide a detailed analysis of the NMS evaluation on ScanNet25k~\cite{gumeli2022roca}, breaking down per-category translation, rotation, and scale errors for each method. We include only methods with publicly available implementations or released predictions, with the results summarized in~\Cref{tab:breakdown}. SUFLECA's largest advantage lies in translation estimation, where it consistently outperforms the compared supervised methods while achieving comparable rotation and scale errors. The use of metric depth likely contributes useful geometric information for translation estimation; however, depth alone does not explain the observed gains. DINOv3-L~\cite{simeoni2025dinov3} and DUNE-B~\cite{sariyildiz2025dune} use the same depth estimates, yet exhibit substantially larger translation errors, highlighting the importance of SUFLECA's geometrically discriminative feature representation for establishing reliable correspondences.

Across methods, elongated objects such as tables, bookshelves, and cabinets tend to exhibit larger errors. Their full spatial extent can be difficult to infer under partial observations, particularly when large portions of the object are occluded. In addition, repeated structures (e.g., cabinet shelves) and large weakly textured surfaces (e.g., bed side panels or tabletops) can make dense correspondence estimation ambiguous. These observations motivate future work on improving alignment robustness for elongated and weakly textured objects under partial visibility.

\subsection{Variation in accuracy thresholds}
\begin{figure*}[t]
    \centering
    \includegraphics[width=\textwidth]{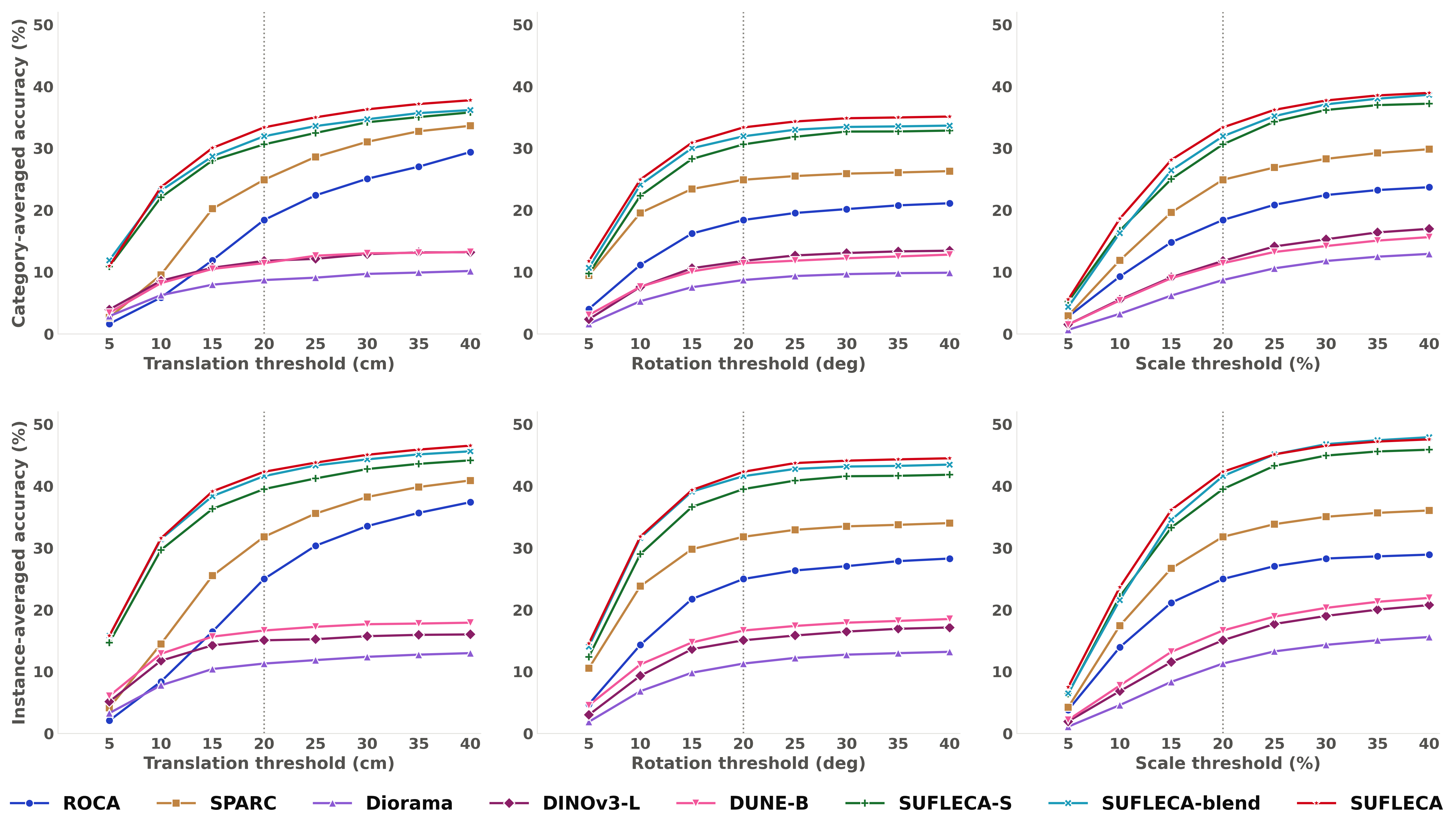}
    \caption{Effect of varying the evaluation thresholds on alignment accuracy on ScanNet25k~\cite{gumeli2022roca}.
    The default criterion is $20\mathrm{cm}$/$20^\circ$/$20\%$ for translation, rotation, and scale, respectively.
    SUFLECA variants outperform the competing methods across all thresholds.
    }
    \label{fig:plots}
\end{figure*}

We additionally evaluate sensitivity to the alignment correctness thresholds in~\Cref{fig:plots}.
For each plot, we vary one threshold over \(5\!-\!40\,\mathrm{cm}\), \(5^\circ\!-\!40^\circ\), or \(5\!-\!40\%\) for translation, rotation, and scale, respectively, while keeping the other two at their default values.
All SUFLECA variants consistently outperform the competing methods across the full range of thresholds, demonstrating that the observed gains are robust to the choice of evaluation threshold.
The performance gap is particularly pronounced under stricter thresholds, indicating that SUFLECA retains its advantage under stricter alignment criteria.

\subsection{Qualitative visualizations}

\begin{figure*}[t]
    \centering
    \includegraphics[width=\textwidth]{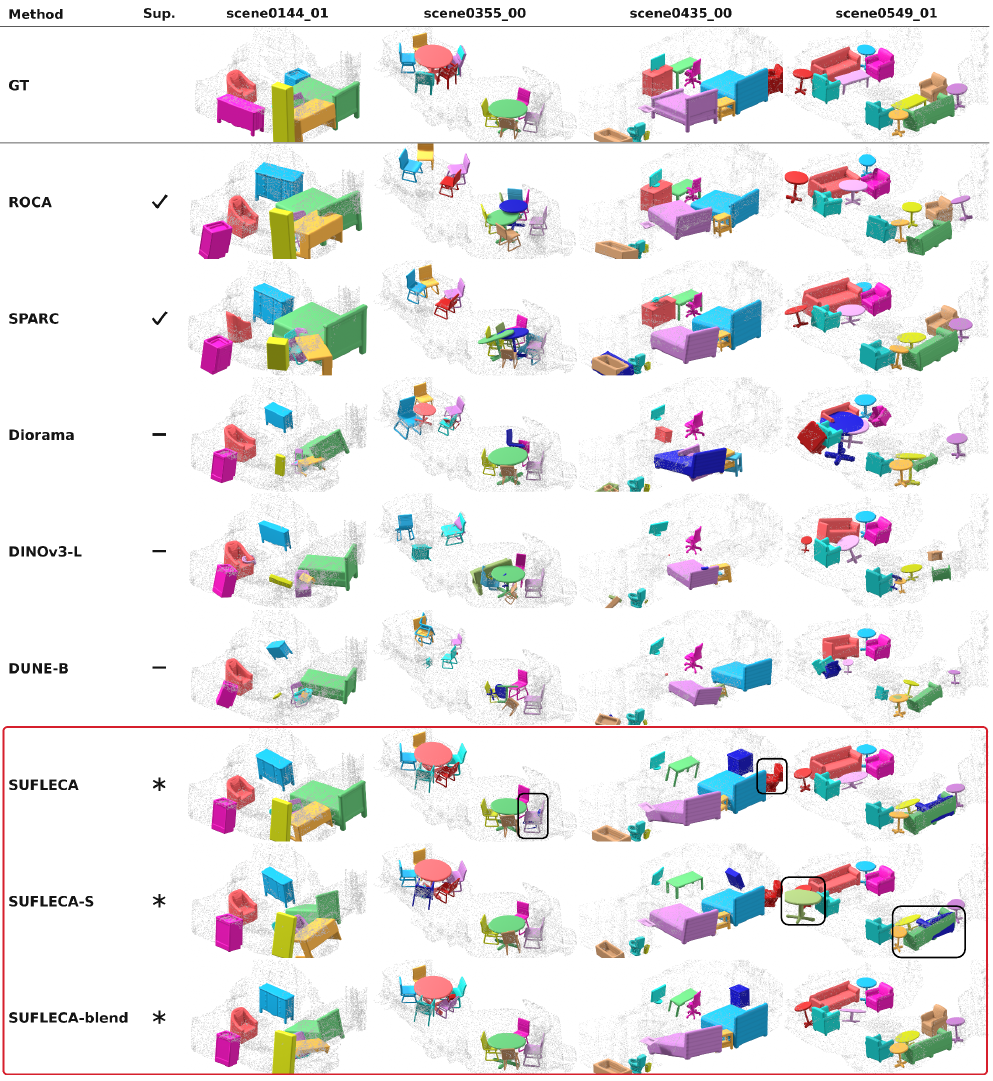}
    \caption{Qualitative comparison of CAD alignments on representative ScanNet25k~\cite{gumeli2022roca} scenes across competing methods.
    Following the standard evaluation protocol, all methods use the bounding boxes and CAD retrievals provided by ROCA~\cite{gumeli2022roca}. As in the main paper, check marks denote supervised methods, dashes denote unsupervised methods, and stars denote weakly supervised methods. Failure cases are highlighted with black boxes.
    From left to right: (1) an undersized alignment remains after NMS because it is a severe outlier; (2) an inexact CAD retrieval matches a toilet seat to a sofa, for which SUFLECA and SUFLECA-S produce plausible alignments but SUFLECA-blend fails; (3) the rotational symmetry of a table introduces ambiguous correspondences and noisy alignments that are not suppressed; and (4) elongated objects that are never fully visible within a single frame are incorrectly reconstructed as multiple instances.
    }
    \label{fig:scene_grid1}
\end{figure*}

\begin{figure*}[t]
    \centering
    \includegraphics[width=\textwidth]{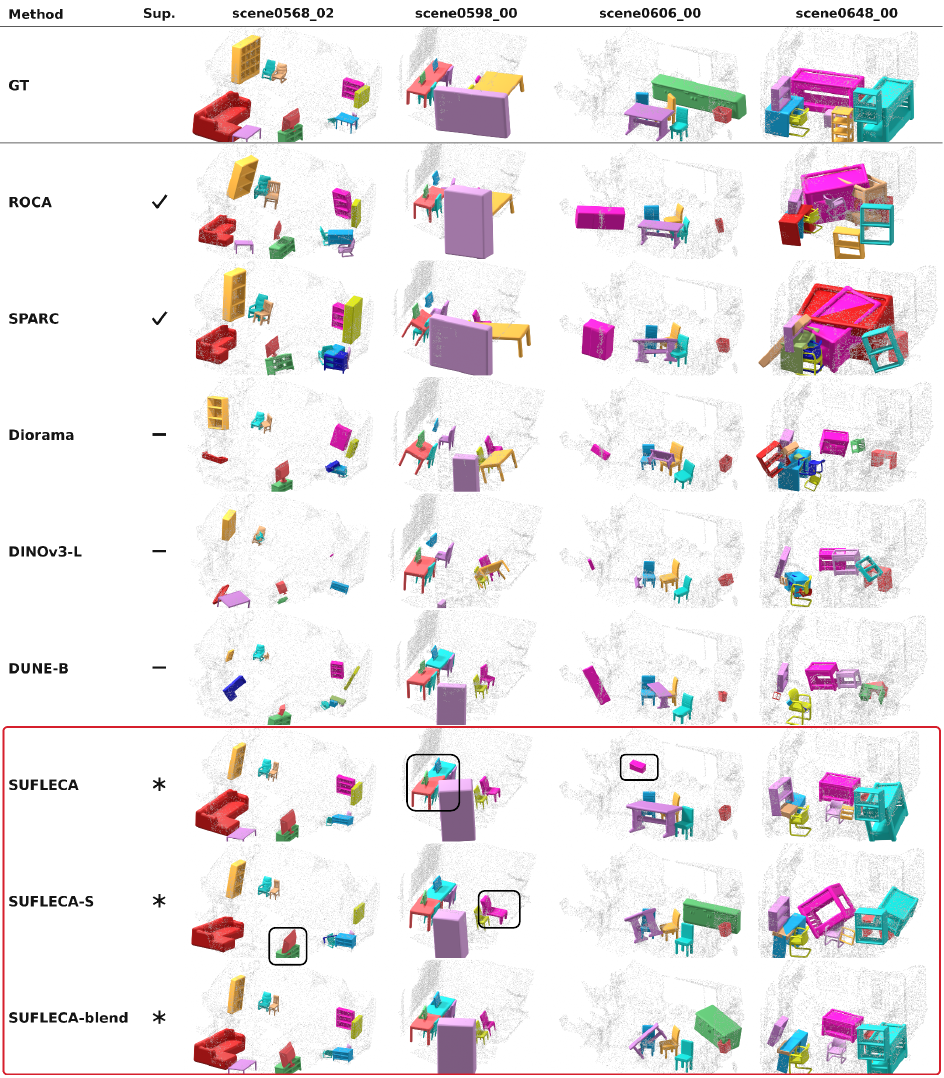}
    \caption{Additional qualitative comparisons on representative ScanNet25k~\cite{gumeli2022roca} scenes, using the same evaluation protocol and visualization as~\Cref{fig:scene_grid1}. Failure cases are highlighted with black boxes.
    From left to right: (1) the absence of scene-level coherence allows independently aligned CAD models to interpenetrate; (2) elongated objects are inherently ambiguous under partial observations and may therefore be reconstructed as multiple instances; (3) large, weakly textured tables are difficult to align and can occlude chair seats beneath them, making the chairs' dimensions along the hidden axes ambiguous; and (4) some visible objects lack CAD annotations, as illustrated by the fuchsia-colored object absent from the ground truth.}
    \label{fig:scene_grid2}
\end{figure*}

\begin{figure*}[t]
    \centering
    \includegraphics[width=\textwidth]{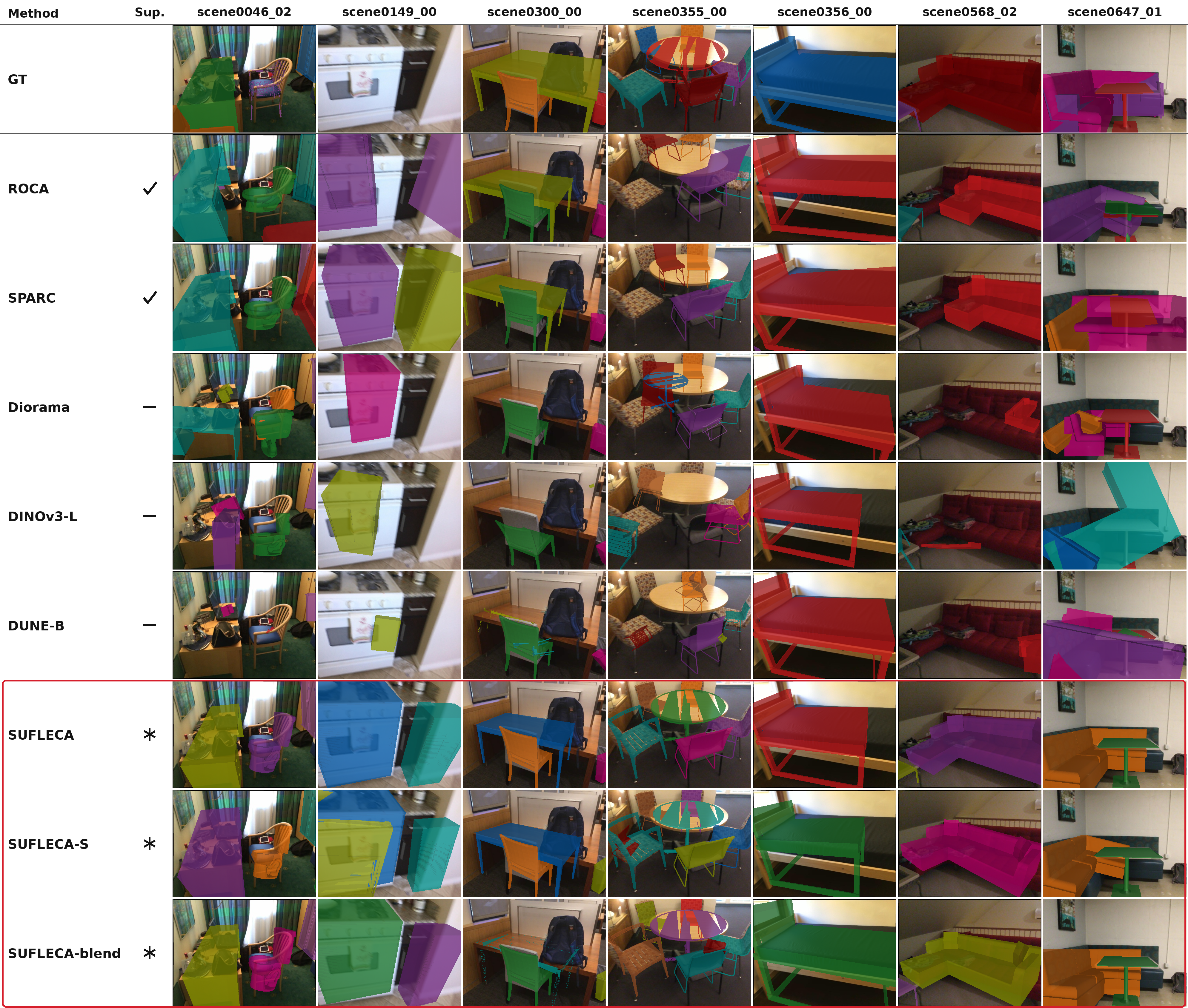}
    \caption{Qualitative comparison of CAD-to-image alignment results on representative ScanNet25k~\cite{gumeli2022roca} images. Retrieved CAD models are overlaid onto the RGB observations using the estimated 9D alignments. Following the standard evaluation protocol, all methods use bounding boxes and CAD retrievals provided by ROCA~\cite{gumeli2022roca}.}
    \label{fig:image_grid}
\end{figure*}

We also visualize the alignments produced by SUFLECA and the competing methods, both in 3D in~\Cref{fig:scene_grid1,fig:scene_grid2} and overlaid on the input RGB images in~\Cref{fig:image_grid}, using examples from the ScanNet25k~\cite{gumeli2022roca} validation split.
We select representative scenes illustrating both successful alignments and characteristic failure cases, which are highlighted with black boxes and described in the corresponding captions of~\Cref{fig:scene_grid1,fig:scene_grid2}.
For scene-level evaluation, we follow the standard non-maximum suppression (NMS)-based protocol introduced by Vid2CAD~\cite{vid2cad}.
While the quantitative metrics measure alignment accuracy, these visualizations additionally reveal spurious alignments.
Compared with the competing zero-shot baselines, SUFLECA produces more accurate and complete scene reconstructions, with fewer spurious and missing alignments.
Common failure cases include inexact CAD retrieval, rotational symmetries, elongated objects under partial visibility, weakly textured surfaces, and objects without corresponding CAD annotations.

\clearpage
\clearpage
{
    \small
    \bibliographystyle{ieeenat_fullname}
    \bibliography{main}
}

\end{document}